\documentclass{article}
\usepackage[accepted]{icml2026}
\IfFileExists{stfloats.sty}{\usepackage{stfloats}}{}
\usepackage{microtype}
\usepackage{graphicx}
\usepackage{subcaption}
\usepackage{booktabs} 

\usepackage{xspace}
\usepackage{xcolor}
\usepackage{adjustbox}
\usepackage{array}
\usepackage{float}
\usepackage{setspace}
\usepackage{multirow}
\usepackage{longtable}
\usepackage{tabularx}
\usepackage{xltabular}
\usepackage{threeparttable}
\usepackage{siunitx}

\usepackage{amsmath}
\usepackage{amssymb}
\usepackage{amsfonts}
\usepackage{mathtools}
\usepackage{bm}
\usepackage{dsfont}
\usepackage{enumitem}
\usepackage{tikz}
\usepackage{pgfplots}
\pgfplotsset{compat=1.18}
\usepackage{listings}
\usepackage[table]{xcolor}
\usepackage{fontawesome5}
\usepackage[most]{tcolorbox}
\usepackage{CJKutf8}
\usepackage{lipsum}
\usepackage{wrapfig}
\definecolor{darkgreen}{HTML}{006400}
\definecolor{mydarkblue}{rgb}{0,0.08,0.45}
\newtcolorbox[auto counter, number within=section, list type=subsubsection, list inside=toc]{sectionbox}[2][]{
colback=white!98!gray, colframe=black, 
colbacktitle=white!90!gray, coltitle=black, 
fonttitle=\bfseries,
title={#2}, 
list entry={Comment \thetcbcounter\quad}
}

\usepackage{hyperref}
\hypersetup{
    colorlinks=true,
    linkcolor=mydarkblue,
    citecolor=mydarkblue, 
    urlcolor=mydarkblue,
}

\usepackage{icml2026}

\usepackage{amsthm}

\usepackage[capitalize,noabbrev]{cleveref}

\theoremstyle{plain}

\theoremstyle{definition}

\theoremstyle{remark}

\newcommand{\method}{\textbf{MARS-SQL}}
\newcommand{\myicon}{\includegraphics[height=1.5em]{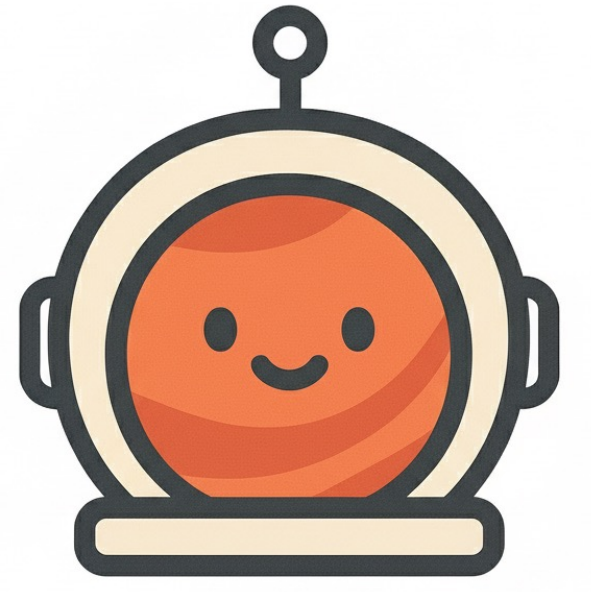}}

\icmltitlerunning{MARS-SQL: A Multi-Agent Reinforcement Learning Framework For Text-To-SQL}

\begin{document}
\twocolumn[
  \icmltitle{\raisebox{-0.3\height}\myicon \method: A Multi-Agent Reinforcement Learning Framework\\For Text-To-SQL}



  \icmlsetsymbol{equal}{*}
  \begin{icmlauthorlist}
    \icmlauthor{Haolin Yang}{equal,hkust}
    \icmlauthor{Jipeng Zhang}{equal,hkust}
    \icmlauthor{Zhitao He}{hkust}
    \icmlauthor{Alexander Zhou}{polyu}
    \icmlauthor{Yi R. (May) Fung}{hkust}
  \end{icmlauthorlist}

  \icmlaffiliation{hkust}{Hong Kong University of Science and Technology}
  \icmlaffiliation{polyu}{Hong Kong Polytechnic
  University}

  \icmlcorrespondingauthor{Yi R. (May) Fung}{yrfung@ust.hk} 


  \vskip 0.3in
]


\printAffiliationsAndNotice{} 

\begin{abstract}
Large Language Models (LLMs) often struggle with the precise logic and schema alignment required for complex Text-to-SQL tasks. While current methods rely heavily on static prompting, they lack the ability to dynamically adapt and self-correct through environmental interaction. To bridge this gap, we propose \textbf{MARS-SQL}, a trainable multi-agent framework for Text-to-SQL. Rather than introducing a new standalone SQL primitive, \textbf{MARS-SQL} makes an agentic workflow trainable by decomposing the problem into three specialized roles: schema grounding, query generation, and solution validation. Central to our approach is a generation agent trained via a multi-turn RL policy within a ReAct-style loop. The agent learns to iteratively reason, execute intermediate SQL actions on a live database, and refine its strategy based on execution feedback. To improve robustness, we further introduce a validation mechanism that treats solution selection as a generative modeling task, identifying the optimal interaction trajectory through next-token prediction probabilities. Empirical evaluations demonstrate the effectiveness of coupling interactive learning with trajectory ranking. \textbf{MARS-SQL} achieves state-of-the-art performance, recording an execution accuracy of 77.84\% on the BIRD development dataset and 89.75\% on the Spider test dataset, while also transferring strongly to out-of-domain benchmarks.
Code is available at \url{https://github.com/YangHaolin0526/MARS-SQL}.
\end{abstract}

\section{Introduction}
Translating natural language questions into executable Structured Query Language (SQL) is an essential task that allows non-expert users to access structured data~\citep{xie2025sdesqlenhancingtexttosqlgeneration,li_dawn_2024,li2023llmservedatabaseinterface}. Recent Large Language Models (LLMs) can generate simple queries for well-organised academic benchmarks. However, they often struggle with the complexity of real-world enterprise databases~\citep{hong2025nextgenerationdatabaseinterfacessurvey,lei2025spider20evaluatinglanguage}. To bridge this gap and tackle the challenges of interacting with complex, real-world databases, researchers have started developing SQL agents~\citep{li_sql-factory_2025,wang_mac-sql_2025,li_swe-sql_2025}. Instead of producing a query in one step, an SQL agent allows an LLM to interact with the database through multiple rounds of reasoning and feedback. This interactive process resembles how human analysts explore data, making it a more natural and effective way to handle complex database tasks.

Current methodologies in the broader field of AI agents have explored several distinct avenues. A prominent strategy is the use of multi-agent systems, where a complex task is decomposed into specialized sub-tasks, each handled by a dedicated agent~\citep{chang2025main,huang_resilience_2025,liu2025costbench,huang2025mac}. A parallel line of work uses test-time scaling methods that generate multiple candidate queries and then select the best one~\citep{ni_lever_2023,li_competition-level_2022}. In the specific domain of Text-to-SQL, these methodologies manifest in two primary forms. One approach relies on monolithic models, which handle schema comprehension, logical planning, and SQL generation in a single pass~\citep{pourreza_reasoning-sql_2025,li_codes_2024}. Another prominent approach involves multi-agent frameworks that improve modularity by using API calls to closed-source LLMs, where different agent roles are defined mainly through prompting~\citep{pourreza2024chasesqlmultipathreasoningpreference,liu_xiyan-sql_2025,xie2025leveraging}.

\textbf{MARS-SQL} addresses three main challenges facing SQL agents: (i) \textbf{Compositional reasoning:} Agents often struggle to formulate and maintain a coherent long-term plan required for complex queries. For example, they may fail to correctly combine multiple SQL clauses like joins and subqueries. This often results in a loop of fixing minor syntax without ever addressing the flawed high-level logic.~\citep{chaturvedi_sql--thought_2025}. To tackle this, we introduce a \textbf{Generation Agent} trained via interactive reinforcement learning, which adopts a multi-turn reasoning process to decompose complex goals and refine the query step-by-step.
(ii) \textbf{Schema understanding:} Navigating schemas with massive amounts of irrelevant tables and columns (distractors) is prone to error. An agent might, for instance, hallucinate columns or misidentify join keys due to the large search space, leading to unproductive interactions~\citep{deng2025reforce}. We address this via a specialised \textbf{Grounding Agent} that performs reasoning-driven schema identification, filtering out distractors to provide a precise schema context for generation. This role separation is especially helpful when schema grounding and SQL generation would otherwise compete for the same context budget. (iii) \textbf{Environmental grounding:} While interactivity is central to agents, current models often lack the nuanced ability to fully leverage environmental feedback~\citep{huang2025scaling}. They struggle to diagnose specific SQL dialect errors or recover from ambiguous execution outcomes, limiting their self-correction capabilities~\citep{zhang2025exesql}. \textbf{MARS-SQL} addresses this via a dual strategy: a \textbf{Generation Agent} that leverages continuous interaction to decompose complex problems and execute intermediate queries to guide its reasoning, and a \textbf{Validation Agent} that executes and verifies the final trajectories against the environment.

\begin{figure}[t]
    \centering
    \includegraphics[width=0.85\linewidth]{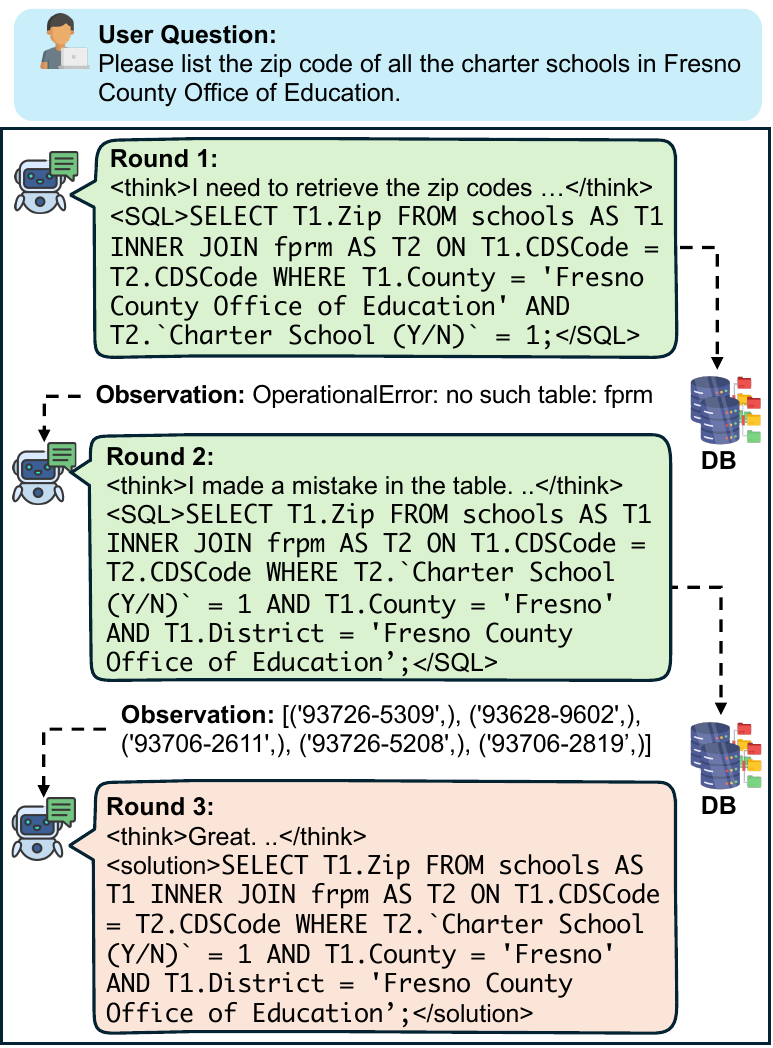}
    \caption{Illustration of interactive reasoning process.}
    \vspace{-0.4cm}
    \label{fig:illustration}
\end{figure}

While recent multi-agent frameworks have explored decomposing Text-to-SQL into modular stages~\citep{gorti2025msc,liu_xiyan-sql_2025}, these approaches primarily rely on static prompting or supervised fine-tuning of distinct modules. \textbf{MARS-SQL} distinguishes itself by introducing a novel interactive training paradigm. Rather than relying on architectural decomposition alone, we empower the Generation Agent with a multi-turn Reinforcement Learning (RL) policy trained via Group Relative Policy Optimization (GRPO) with sparse execution rewards. This allows the agent to learn how to navigate the schema and self-correct errors dynamically, rather than merely following a prompted chain of thought. 
The core innovation lies in the Generation Agent's ability to operate within a ReAct-style Think-Act-Observe loop~\citep{yao2022react}. As detailed in Figure~\ref{fig:illustration}, this process transforms SQL generation into a dynamic investigation. For instance, when the agent initially attempts to join the hallucinated ``fprm'' table, it encounters an \texttt{OperationalError}. Rather than failing silently, the agent explicitly interprets this feedback, diagnoses the schema error, and self-corrects to identify the correct ``frpm'' table.
To further improve robustness, multiple reasoning trajectories are explored during inference, and the Validation Agent selects the optimal result. Our deployed system uses three specialized 7B agents, corresponding to an inference-time footprint of approximately 21B parameters. Through this synergistic workflow, \textbf{MARS-SQL} achieves strong empirical results, reaching a new state-of-the-art Execution Accuracy of 77.84\% on the BIRD development set and 89.75\% on the Spider test set. Even when compared through efficiency-oriented analyses under matched latency budgets (Appendix~\ref{sec:appendix_efficiency}), MARS-SQL outperforms strictly prompted or self-consistency baselines, supporting the view that its gains are not explained by added parameters or inference time alone.

Our main contributions are as follows:
\begin{enumerate}
    \item We introduce \textbf{MARS-SQL}, a trainable multi-agent framework for Text-to-SQL. Our contribution is primarily at the workflow level: we separate schema grounding, interactive generation, and outcome validation, and train these roles instead of relying on a static prompted pipeline.
    \item We show how to train an \textbf{interactive multi-turn generation policy} with execution feedback. The Generation Agent operates within a Think--Act--Observe loop and learns to diagnose errors and self-correct from live database feedback, while the Grounding Agent further improves schema context management through data-driven training.
    \item \textbf{MARS-SQL} achieves strong empirical results, reaching 77.84\% execution accuracy on BIRD-dev and 89.75\% on Spider-test, while also generalizing well to additional out-of-domain benchmarks. We further show that the gains are not reducible to raw parameter scaling alone, and provide an efficiency analysis that characterizes the accuracy--latency--memory trade-off of the full framework rather than evaluating it only as a single operating point.
\end{enumerate}

\section{Preliminaries}
\textbf{Background Formulation.} The primary goal of a Text-to-SQL system is to translate a natural language question into an executable SQL query. We can formally define this task as learning a mapping from a user question and a group of database schemas to the corresponding SQL query.

\begin{figure*}[t]
    \centering
    \includegraphics[width=0.95\textwidth]{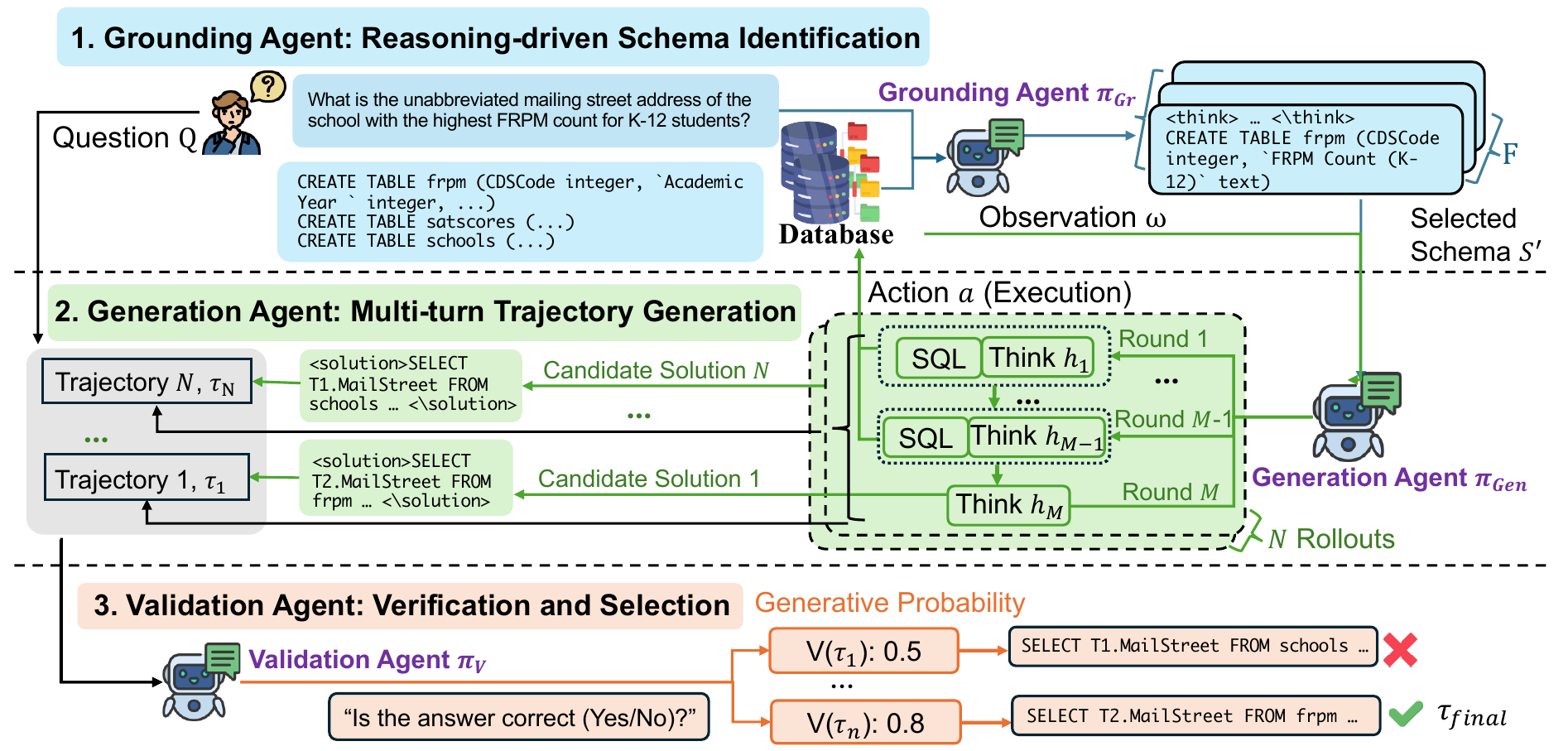}
    \caption{The three-stage workflow of MARS-SQL. (1) Grounding: A Grounding Agent selects the relevant schema. (2) Generation: A Generation agent produces multiple interaction trajectories using a Think-Act-Observe loop. (3) Validation: A Validation agent scores and selects the best trajectory.}
    \vspace{-2pt}
    \label{fig:motivation}
\end{figure*}
Let $Q$ be the natural language question posed by a user. Let $S$ be the database schema, which defines the structure of the database. The schema $S$ consists of a set of tables $T = \{t_1, t_2, \dots, t_m\}$, where each table $t_i$ is composed of a set of columns $C_i = \{c_{i,1}, c_{i,2}, \dots, c_{i,k}\}$. The schema also includes information about data types, primary keys (PKs), and foreign keys (FKs) that define the relationships between tables. The objective is to generate a SQL query $Y$ such that when it is executed on the database instance $D$, it produces the correct answer to the question $Q$.

Conventionally, the Text-to-SQL problem is treated as a sequence-to-sequence translation task, where the goal is to learn a function $f$:
Y = f(Q, S)
This formulation, however, treats the generation as a single, static step and fails to capture the exploratory and corrective nature required for solving complex analytical queries.

\textbf{Reformulation as an Interactive Decision Process.}
As highlighted in the introduction, the static, one-shot formulation is insufficient for complex reasoning. A human analyst does not simply translate; they interact, explore, and refine. To model this more robust process, we reformulate Text-to-SQL as a sequential decision-making task, grounded in the ReAct paradigm~\citep{yao2023reactsynergizingreasoningacting}.

Instead of learning a direct mapping to a final query, our goal is to learn an optimal \textbf{policy}, $\pi$, that generates a \textbf{trajectory} of thoughts and actions to solve the problem. A complete interaction trajectory, $\tau$, is a sequence of multiple rounds:
\begin{equation}
    \tau = (h_1, \alpha_1, \omega_1, \dots, h_M, \alpha_M, \omega_M)
\end{equation}
Each turn in the trajectory consists of:
\begin{itemize}
    \item \textbf{Thought ($h_t$):} An internal reasoning step where the agent analyzes the problem state, reflects on past observations, and plans the next action.
    \item \textbf{Action ($\alpha_t$):} An operation chosen by the agent from a predefined action space $\mathcal{A}$, primarily involving the execution of SQL queries against the database.
    \item \textbf{Observation ($\omega_t$):} The feedback received from the environment after executing action $\alpha_t$. This could be a query result, a database error, or other information that guides the agent's next thought.
\end{itemize}

Under this formulation, the objective is to learn an optimal policy $\pi(\alpha_t | Q, S, (h_{<t}, \alpha_{<t}, \omega_{<t}))$ that maximizes the expected total reward over the trajectory, $E[R(\tau)]$. The reward $R(\tau)$ is typically determined by the final outcome, specifically, whether the trajectory successfully produces a correct SQL query. This interactive, policy-based formulation naturally accommodates the trial-and-error and self-correction that are essential for tackling complex, real-world database queries.
\section{Methodology}
As illustrated in Figure \ref{fig:motivation}, we introduce \method, a trainable multi-agent framework that treats Text-to-SQL generation as an interactive, tool-augmented decision-making process. Our goal is not to propose a new standalone SQL primitive, but to make role decomposition and agent--database interaction trainable. The framework operates in three stages: Grounding, Generation, and Validation. Initially, a Grounding Agent prunes the full database schema to only the tables and columns relevant to the user question. Subsequently, a Generation agent executes a multi-turn rollout, producing multiple distinct interaction trajectories by actively querying the database. Finally, a Validation Agent scores each trajectory, and the one with the highest confidence score is selected as the final answer. 

\subsection{Grounding Agent: Reasoning-driven Schema Identification}
The Grounding Agent performs table-level schema linking. Its goal is to learn a policy $\pi_{Ground}$. For each table $t_i \in T( 1 \leq i \leq F)$ and the user's question $Q$, the agent takes the pair $x_i = (Q, t_i)$ as input. It then generates a structured output $o_i = (d_i, C'_i)$, where $d_i \in \{\text{`Y', `N'}\}$ is the relevance decision and $C'_i \subseteq C_i$ is the predicted subset of essential columns. The final output of this stage is the reduced schema $S'$, containing only the tables and columns deemed relevant: $S' = \{(t_i, C'_i) \mid o_i \text{ has } d_i = \text{`Y'}\}$.\\

\textbf{Training Algorithm. }
We train the agent using \textbf{Group Relative Policy Optimization (GRPO)}~\citep{shao_deepseekmath_2024}. For each input $x_i$, the model generates a group of $G$ candidate outputs $\{o_1, \dots, o_G\}$. The policy $\pi_\theta$ is then updated via the objective:
\begin{equation}
\label{eq:grpo_grounding}
\small 
\begin{split}
    J_{\text{GRPO}}(\theta) = \mathbb{E} \Bigg [ \frac{1}{G} \sum_{j=1}^{G} \min \bigg( \frac{\pi_\theta(o_j|x_i)}{\pi_{\theta_{\text{old}}}(o_j|x_i)} A_j, \\
    & \hspace{-17.5em} \text{clip} \left( \frac{\pi_\theta(o_j|x_i)}{\pi_{\theta_{\text{old}}}(o_j|x_i)}, 1-\epsilon, 1+\epsilon \right) A_j \bigg) - \beta D_{\text{KL}}(\pi_\theta \| \pi_{\text{ref}}) \Bigg]
\end{split}
\end{equation}
where $A_j$ is the advantage for candidate $o_j$. The agent's prompt template is in Appendix~\ref{tab:prompt_schema_linking}.

\textbf{Reward Design.}
The reward function $R_{Ground}$ provides a granular score based on the accuracy of the agent's prediction. Let the agent's parsed prediction be $P = (d_p, C_p)$, where $d_p \in \{\text{`Y', `N'}\}$ is the relevance decision and $C_p$ is the set of predicted columns. Let the ground truth be $o^* = (d_g, C_g)$. The reward $R_g(o, o^*)$ is defined as:
\[
R =
\begin{cases}
1.0 & \text{if } o = o^*  \\
\max(0.5, \frac{|C_g|}{|C_p|}) & \text{if } d_p = d_g = \text{`Y'} \text{ and } C_g \subset C_p  \\
0.2 & \text{if } d_p = \text{`Y'} \text{ and } d_g = \text{`N'}  \\
0.1 & \text{if } d_p = d_g = \text{`Y'} \text{ and } C_g \not\subseteq C_p  \\
0.0 & \text{if response format is invalid}
\end{cases}
\]
This heuristic reward shaping is grounded in the principle that missing schema elements (False Negatives) are fatal, while including extra ones (False Positives) is merely noisy.

\subsection{Generation Agent: Multi-turn Trajectory Generation}
The Generation Agent is the central component, tasked with producing SQL queries. Its \textbf{input} is the user question $Q$ and the reduced schema $S'$ from the Grounding Agent. Its \textbf{output} is a set of $N$ candidate interaction trajectories, $\{\tau_1, \dots, \tau_{N}\}$, where each trajectory comprises of $M$ rounds of the Think-Act-Observe process. The correct trajectory is expected to result in the final SQL solution $Y_i$.

\textbf{MDP Formulation. }We model the multi-turn generation process as a Markov Decision Process (MDP), defined by the tuple $(\mathcal{S}, \mathcal{A}, P, R)$.
\begin{itemize}
    \item \textbf{State Space $\mathcal{S}$}: A state $s_t$ represents the history of interaction up to round $t$, containing the sequence of past thoughts, actions, and observations $((h_1, \alpha_1, \omega_1), \dots, (h_{t-1}, \alpha_{t-1}, \omega_{t-1}))$.
    \item \textbf{Action Space $\mathcal{A}$}: An action $a_t = (h_t, \alpha_t)$ consists of generating a thought $h_t$ and an executable SQL $\alpha_t$.
    \item \textbf{Transition $P$}: $P(s_{t+1}|s_t, a_t)$ is the transition probability, which is determined by the environment (i.e., the database executing the action $\alpha_t$).
    \item \textbf{Reward $R$}: The reward function $R_{gen}(\tau)$ provides a sparse signal based on the final outcome of a complete trajectory $\tau$.
\end{itemize}
The goal is to learn a policy $\pi_{Gen}(a_t|s_t)$ that maximizes the return $J(\pi_{Gen}) = \mathbb{E}_{\tau \sim \pi_{Gen}}[R_{Gen}(\mathcal{T})]$.

\textbf{Training. }We train the policy $\pi_{Gen}$ using Group Relative Policy Optimization (GRPO). For an input $(Q, S')$, we generate a group of $G$ trajectories $\{\tau_1, \dots, \tau_G\}$, where each trajectory $tau_i$ consists of a sequence of states and actions $(s_0^i, a_0^i, s_1^i, \dots)$. The GRPO objective for trajectories is defined as formula \ref{eq:grpo_generation},
\begin{equation}
\label{eq:grpo_generation}
\makebox[\columnwidth][l]{%
\scalebox{0.79}{%
  $\displaystyle
\begin{split}
J_{\text{GRPO}}(\theta) = \mathbb{E}_{\substack{(Q,S') \sim \mathcal{D}, \\ \{\tau_i\}_{i=1}^G \sim \pi_{\theta_{\text{old}}}}} \Bigg[ \frac{1}{G} \sum_{i=1}^{G} \sum_{t=0}^{|\tau_i|-1} \sum_{j=1}^{|a_t^i|} \min \bigg( \frac{\pi_\theta(a_{t,j}^i|s_t^i, a_{t,<j}^i)}{\pi_{\theta_{\text{old}}}(a_{t,j}^i|s_t^i, a_{t,<j}^i)} A_i, \\
& \hspace{-26.5em} \text{clip} \left( \frac{\pi_\theta(a_{t,j}^i|s_t^i, a_{t,<j}^i)}{\pi_{\theta_{\text{old}}}(a_{t,j}^i|s_t^i, a_{t,<j}^i)}, 1-\epsilon, 1+\epsilon \right) A_i \bigg) \Bigg]
\end{split}
$%
}%
}
\end{equation}
where $a_{t,j}^i$ is the $j$-th token of action $a_t^i$ in trajectory $\tau_i$, and $A_i$ is the advantage for the entire trajectory, computed based on the relative rewards of all trajectories within the group. The reward signal $R_{gen}(\tau)$ used to compute $A_i$ is derived solely from execution outcomes, encouraging the agent to prioritize both syntactic validity and semantic correctness:
\[
R_{gen}(\tau) =
\begin{cases}
1.0 & \text{if valid and execution correct} \\
0.0 & \text{if valid but incorrect} \\
-1.0 & \text{if invalid}
\end{cases}
\]
This coarse but decisive feedback gives the agent freedom to discover effective reasoning strategies without being constrained to annotated step-level traces. Unlike Supervised Fine-Tuning (SFT) which typically relies on expensive, human-annotated Chain-of-Thought data, our sparse reward mechanism allows the agent to learn directly from environmental feedback. This enables the agent to autonomously discover diverse and robust error-recovery strategies that are often absent in static training sets. Consequently, MARS-SQL demonstrates superior data efficiency, achieving state-of-the-art results with only $\sim$35k training instances, whereas SFT baselines often require millions of synthetic examples to adequately cover the error space (detaile in Appendix~\ref{sec:training_cost}).

\textbf{Interactive Reasoning.} The agent is grounded in the ReAct paradigm, interleaving reasoning and acting in a Think-Act-Observe loop. This iterative structure transforms SQL generation from a one-shot translation into a dialogue with the database, enabling robust recovery from errors.

\subsection{Validation Agent: Verification and Selection}
The Validation Agent selects the optimal solution from the multiple candidates generated. Its \textbf{input} is the set of $N$ candidate trajectories $\{\tau_1, \dots, \tau_N\}$ and the original question $Q$. Its \textbf{output} is the single best trajectory, $\tau_{\text{final}}$. We employ a Generative Validation agent $V$, reframing verification as a next-token prediction task that leverages the base model's own capabilities.

\textbf{Training and Inference. }
The Validation Agent is trained via supervised fine-tuning (SFT) to generate a single token response: ``Yes'' for a correct trajectory or ``No'' for an incorrect one, conditioned on the question and trajectory. The prompt structure is in Appendix~\ref{Training Dataset}.

At inference time, the agent's score for a trajectory $\tau_i$ is the average probability of the ``Yes'' token across  
$M$ stochastic reasoning rounds :
\begin{equation}
V(\tau_i) = \frac{1}{M} \sum_{j=1}^{M} P(y_j = \text{``Yes''}| \tau_i, Q)
\end{equation}
The trajectory with the highest confidence score is selected as the final answer.
This method effectively turns the generative model into a high-quality reranker, capable of discerning the most plausible and accurate reasoning path among many alternatives.

\section{Experiment}
\subsection{Experiment Setup}
\textbf{Implementations \& Efficiency} 

All agents are based on Qwen2.5-Coder-7B-Instruct, implemented in PyTorch and trained on NVIDIA H800 GPUs. The Grounding and Generation Agents utilize RL via the Verl~\citep{sheng2024hybridflow} and SkyRL~\citep{liu2025skyrlsql} frameworks, respectively, while the Validation Agent is fine-tuned (SFT) on agent-generated trajectories. 
For inference, we adopt $G=8$ rollouts and $M=8$ verification rounds. This is our performance-oriented configuration and delivers the best overall accuracy. Since the deployed system contains three specialized 7B agents, we explicitly report both the per-agent size (7B) and the total inference-time footprint ($\sim$21B). We analyze its latency, token cost, VRAM footprint, normalized-budget comparisons, and alternative lower-cost operating points in Appendix~\ref{sec:appendix_efficiency}. The entire multi-agent training pipeline requires only $\sim$13 hours (Appendix~\ref{sec:training_cost}). Detailed prompts and hyperparameters are provided in Appendix~\ref{grounding},~\ref{Multi-turn Generation} and~\ref{Training details}.


\textbf{Benchmark Dataset }

All experiments are conducted on the BIRD~\citep{li2023llmservedatabaseinterface}, Spider 1.0~\citep{yu2019spiderlargescalehumanlabeleddataset} and Spider-DK~\citep{gan2021exploringunderexploredlimitationscrossdomain} datasets. We use BIRD for in-domain evaluation and Spider and Spider-DK for out-of-domain evaluation; we further report SParC results in Appendix~\ref{sec:additional_ood}. Dataset details are in Appendix~\ref{Evaluation Dataset}.

\begin{table*}[ht]
\centering
\caption{Main results on the BIRD-dev, Spider-test, and Spider-DK benchmarks. We report Execution Accuracy (\%). `Thinking?' indicates whether the method uses a multi-step reasoning process. For multi-agent systems, the size column reports the per-agent model size; \method uses three 7B agents ($\sim$21B total at inference). \textbf{Bold} indicates the best result, and \underline{underline} indicates the second best.}
\label{tab:main_results}
\resizebox{\textwidth}{!}{
\begin{tabular}{@{}llcccccc@{}}
\toprule
\textbf{Model} & \textbf{Size} & \textbf{\# agents} & \textbf{Thinking?} & \textbf{Training set} & \textbf{Bird-dev (\%)} & \textbf{Spider-test (\%)} & \textbf{Spider-DK (\%)} \\
\midrule
\multicolumn{7}{c}{\textit{Base Models}} \\
\midrule
O3-mini & - & - & Yes & - & 61.34 & 78.82 & 71.77 \\
Qwen-2.5-coder & 7B & - & No & - & 54.56 & 75.87 & 61.31 \\
Qwen3-coder & 30B & - & No & - & 59.00 & 79.97 & 68.79 \\
GPT-4o & - & - & No & - & 61.90 & 77.10 & 72.9 \\
GPT-5 & - & - & No & - & 65.45 & 78.39 & 66.92 \\
\midrule

\multicolumn{7}{c}{\textit{Closed-source Multi agentic framework}} \\
\midrule
CHESS & - & 4 & No & - & 65.00 & 87.2 & - \\
OpenSearch-SQL+ GPT-4o & - & 4 & No & - & 69.30 & 87.1 & - \\
XiYan-SQL & - & 3 & No & - & 73.34 & \underline{89.65} & - \\
CHASE-SQL + Gemini & - & 4 & Yes & - & \underline{74.90} & 87.6 & - \\
\midrule
\multicolumn{7}{c}{\textit{Open Source Agentic Framework}} \\
\midrule
CodeS & 7B & 1 & No & Spider & 57.17 & 80.3 & 72.0 \\
Share & 8B & 1 & No & Bird & 64.14 & 85.90 & 75.3 \\
OmniSQL & 32B & 1 & No & OmniSQL & 64.5 & 87.60 & 76.1 \\
Arctic-Text2SQL-R1 & 32B & 1 & Yes & Bird+Spider & 70.70 & 88.80 & \textbf{81.5} \\
Reasoning SQL & 14B & 1 & Yes & Bird & 72.29 & 81.43 & 73.03 \\
\textbf{\method} & 7B each & 3 & Yes & Bird & \textbf{77.84} & \textbf{89.75} & \underline{78.13} \\
\bottomrule
\end{tabular}}
\end{table*}
\textbf{Evaluation Metric }

We evaluate model performance using Execution Accuracy (EX), which is the primary metric for correctness. A predicted SQL query receives a score of 1 if its execution result is identical to the execution result of the ground-truth query, and 0 otherwise. The final score is the percentage of correctly executed queries.

\textbf{Baseline models }

To contextualize the performance of \method, we conduct a comprehensive comparison against a diverse set of models. These are organized into three distinct categories: Base models, High-performing closed-source systems, and Trained open-source models.  \\
\textbf{Base Models: } This category includes foundational LLMs used without task-specific fine-tuning to establish a performance baseline. We evaluate O3-mini, GPT-4o~\citep{openai2023gpt4}, GPT-5~\citep{singh2025openai}, Qwen2.5-coder-7B~\citep{hui_qwen25-coder_2024} and Qwen3-coder-30B~\citep{qwen3technicalreport}. These results help gauge the inherent Text-to-SQL capabilities of modern LLMs before specialized training. \\
\textbf{Closed Source Multi agentic framework: } This category consists of systems that leverage powerful proprietary models via APIs, representing the upper bound of performance achievable with leading commercial technology. These methods, such as CHESS ~\citep{talaei_chess_2024}, OpenSearch-SQL ~\citep{xie2025opensearchsqlenhancingtexttosqldynamic}, XiYan-SQL ~\citep{liu_xiyan-sql_2025}, and CHASE-SQL ~\citep{pourreza2024chasesqlmultipathreasoningpreference}, typically employ sophisticated frameworks and prompting techniques. This comparison situates our open-source multi-agent framework performance against industry-leading systems.\\
\textbf{Open Source Agentic Framework: } This group comprises leading open-source models specifically fine-tuned for the Text-to-SQL task, representing the current state-of-the-art in the research community. These models, including CodeS~\citep{li_codes_2024}, Share~\citep{qu2025shareslmbasedhierarchicalaction}, OmniSQL~\citep{li_omnisql_2025}, Arctic-Text2SQL-R1~\citep{yao_arctic-text2sql-r1_2025}, and Reasoning SQL~\citep{pourreza_reasoning-sql_2025}, employ various advanced training methodologies. Comparing \method~against these systems assesses its competitiveness and advancements over existing specialized methods.

\subsection{Main Results}
As presented in Table \ref{tab:main_results}, our method, \method, trained solely on the BIRD training set, achieves state-of-the-art execution accuracy on both the Bird-dev (77.84\%) and the Spider-test (89.75\%). Additionally, it obtains the second-highest score on the Spider-DK benchmark with 78.13\%. 

\textbf{In-Domain Performance on BIRD-dev.}

On the in-domain BIRD-dev set, \textbf{MARS-SQL} establishes a new state-of-the-art with an execution accuracy of \textbf{77.84\%}. This result represents a significant improvement of 5.55\% over the next best open-source competitor, Reasoning SQL (72.29\%). More impressively, our system outperforms all listed closed-source solutions, including the strong CHASE-SQL + Gemini (74.90\%). This demonstrates the superior effectiveness of our training methodology on this complex, real-world benchmark.



\begin{table*}[ht]
\centering
\small
\caption{Ablation study on the components of our multi-agent framework. We evaluate the contribution of each agent (Grounder, Verifier). The final row, \textbf{MARS-SQL}, represents our full proposed model.}
\label{tab:ablation_framework}
\begin{tabular}{@{}llcccc@{}}
\toprule
\textbf{Configuration} & \textbf{Agent Size} & \textbf{\# agents} & \textbf{Bird dev (\%)} & \textbf{Spider test (\%)} & \textbf{Spider DK (\%)} \\
\midrule
Generator Only (Base) & 7B & 1 & 66.37 & 80.11 & 69.91 \\
w/o verifier (Grounder+ RL Generator) & 7B & 2 & 68.71 & 80.72 & 70.65 \\
w/o Grounder (RL Generator + Verifier) & 7B & 2 & 69.75 & 89.19 & 77.01 \\
w/ Self-Consistency (instead of Verifier) & 7B & 2 & 72.93 & 83.51 & 73.08 \\
\midrule
\textbf{\textbf{MARS-SQL} (Full Framework)} & \textbf{7B} & \textbf{3} & \textbf{77.84} & \textbf{89.75} & \textbf{78.13} \\
\bottomrule
\end{tabular}
\end{table*}
\begin{figure*}[h]
    \centering
    \includegraphics[width=\textwidth]{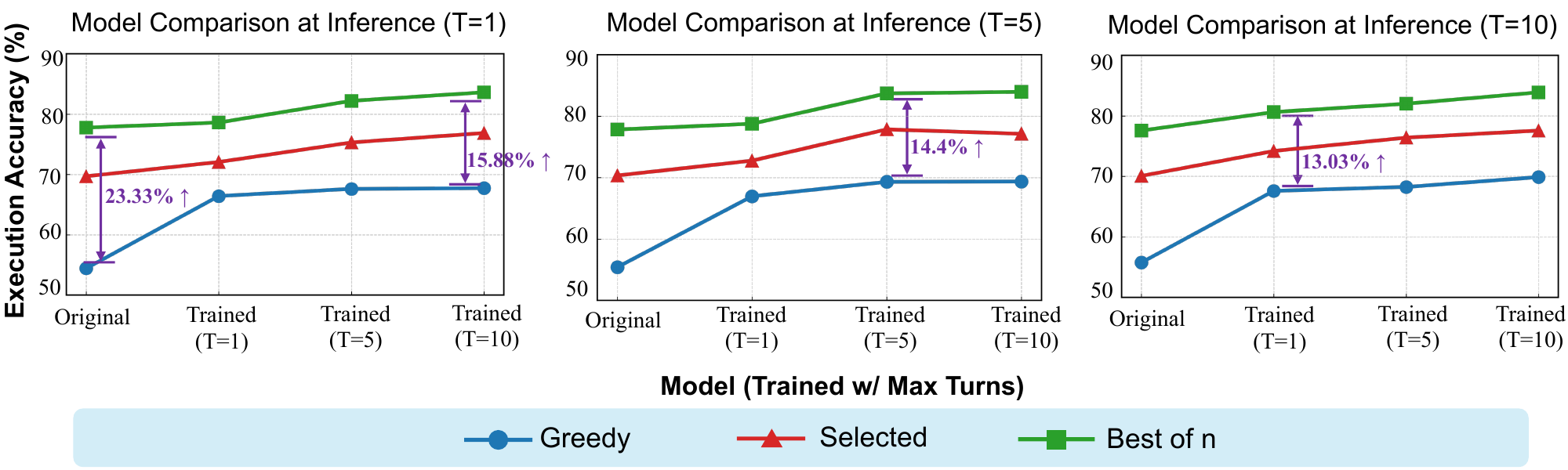}
    \caption{Execution accuracy on Bird-dev of models fine-tuned with different maximum interaction turns (T), evaluated at inference turn limits of 1, 5, and 10. \textbf{`Greedy'} uses a single generation trajectory ($N=1$) without validation; \textbf{`Selected'} denotes the final trajectory chosen by the Validation Agent from $N=8$ candidates; and \textbf{`Best of N'} represents the oracle upper bound where the question is considered correct if any of the $N$ candidates matches.}
    \label{fig:comparison_turns}
\end{figure*}
\textbf{Out-of-Domain Generalization.}

The out-of-domain generalization of \textbf{MARS-SQL} is particularly noteworthy, demonstrated by its strong performance on both the Spider-test and Spider-DK benchmarks. On the broad Spider-test set, it achieves a state-of-the-art score of \textbf{89.75\%}, showcasing exceptional generalization to unseen schemas and question types. This robustness extends to the specialized Spider-DK benchmark, which tests for implicit domain knowledge, where \method secures a competitive second-highest score of \textbf{78.13\%}. Crucially, these results were achieved without any exposure to the Spider training set. This contrasts with competitors like Arctic-Text2SQL-R1, which required training on Spider data (from which Spider-DK is derived) to achieve its high scores. We further include additional out-of-domain evidence on SParC in Appendix~\ref{sec:additional_ood}. Together, these results highlight that training solely on the diverse BIRD dataset effectively equips the model for broad cross-domain and interaction-heavy challenges.

\subsection{Ablation Studies}

\textbf{Multi-agent frame components analysis.}

We conduct a systematic ablation study to validate the contribution of each key component in our \textbf{MARS-SQL} framework, with results presented in Table~\ref{tab:ablation_framework}.
The analysis confirms that both the Schema Grounder and the Generative Validation Agent are useful components, though their impact is not uniform across datasets. The Validation Agent provides the most consistent gain: our purpose-built verifier substantially outperforms a strong alternative like Self-Consistency (77.84\% vs. 72.93\% on BIRD-dev), highlighting the benefits of specialized validation. The Grounder is especially helpful on BIRD-dev and more modest on Spider-test, which is consistent with our later analysis that grounding matters most under larger, noisier, or more ambiguous schemas. We therefore view the framework's gain as arising from complementary roles rather than from every component contributing equally in every setting. This interpretation is further reinforced by single-model multi-task baselines: when a single Qwen2.5-Coder model is SFT-trained to perform grounding, generation, and validation within one shared model, 14B and 32B variants reach only 55.67\% and 57.82\% EX on BIRD-dev, respectively (Appendix~\ref{sec:param_scaling_appendix}).

\textbf{Controlled isolation of component gains.}

To better separate the effects of structured inference, learned interaction, and reranking, we further conduct a dataset-wide controlled-isolation analysis on BIRD-dev. Starting from a strict baseline consisting of the RL Generator only with $T{=}1$ and greedy decoding, we add exactly one mechanism at a time using configurations already studied in our manuscript. The results are summarized in Table~\ref{tab:causal_isolation_main}, with further details in Appendix~\ref{sec:controlled_isolation}.

\begin{table}[h]
\centering
\caption{Dataset-wide controlled isolation on BIRD-dev. We report the additional queries resolved beyond the strict RL-Generator baseline (62.78\% EX).}
\label{tab:causal_isolation_main}

\resizebox{\linewidth}{!}{%
\begin{tabular}{lcc}
\toprule
\textbf{Mechanism} & \textbf{Extra Solved} & \textbf{Gain} \\
\midrule
Schema pruning & 56 & +3.63\% \\
RL interactive reasoning & 44 & +2.89\% \\
Generative verification & 131 & +8.54\% \\
\midrule
\textbf{Total improvement} & \textbf{231} & \textbf{+15.06\%} \\
\bottomrule
\end{tabular}%
}
\end{table}

This analysis sharpens the causal picture in two ways. First, the largest isolated gain comes from generative verification, which is consistent with our ablations and selection-method comparisons. Second, the RL interactive policy still contributes 44 additional correct queries under greedy decoding alone, without relying on sampling or reranking. This provides direct evidence that execution-feedback-driven interaction improves the underlying policy itself rather than only helping through best-of-$N$ selection.


\textbf{Influence of different max interaction turns.}

We then study the impact of the maximum interaction turns (T) during Reinforcement Learning. The results are visualized in Figure~\ref{fig:comparison_turns}, with full details provided in Appendix~\ref{Interaction turns impact}. As shown, increasing T from 1 to 10 consistently improves both Greedy and Best of 8 accuracy. Notably, our model trained with T=10 significantly outperforms models trained with fewer turns across all inference settings. For instance, at Inference (T=10), it achieves 69.88\% Greedy accuracy, surpassing the T=1 model (67.60\%) and the base model (55.76\%). 
Furthermore, this process enhances single-pass reliability by narrowing the gap between Best of 8 (potential) and Greedy (actual) performance. This gap shrinks from a substantial 23.33\% in the base model to 12.19\% in the T=1 model at Inference (T=1). Training with a larger T reinforces this effect, making the model's greedy output more aligned with its optimal potential, thereby improving its dependability. Beyond accuracy, RL also improves interaction efficiency: the trained Generation Agent uses fewer average interaction turns than the base model across difficulty levels, while still retaining the ability to spend more turns on long-tail difficult cases. We report this rollout-length evolution and turn distribution in Appendix~\ref{Multi-turn Generation}, Table~\ref{tab:interaction_turns}, and Table~\ref{tab:turn_distribution}.

\begin{figure}[h]
    \centering
    \vspace{-0.1cm}
    \includegraphics[width=\linewidth]{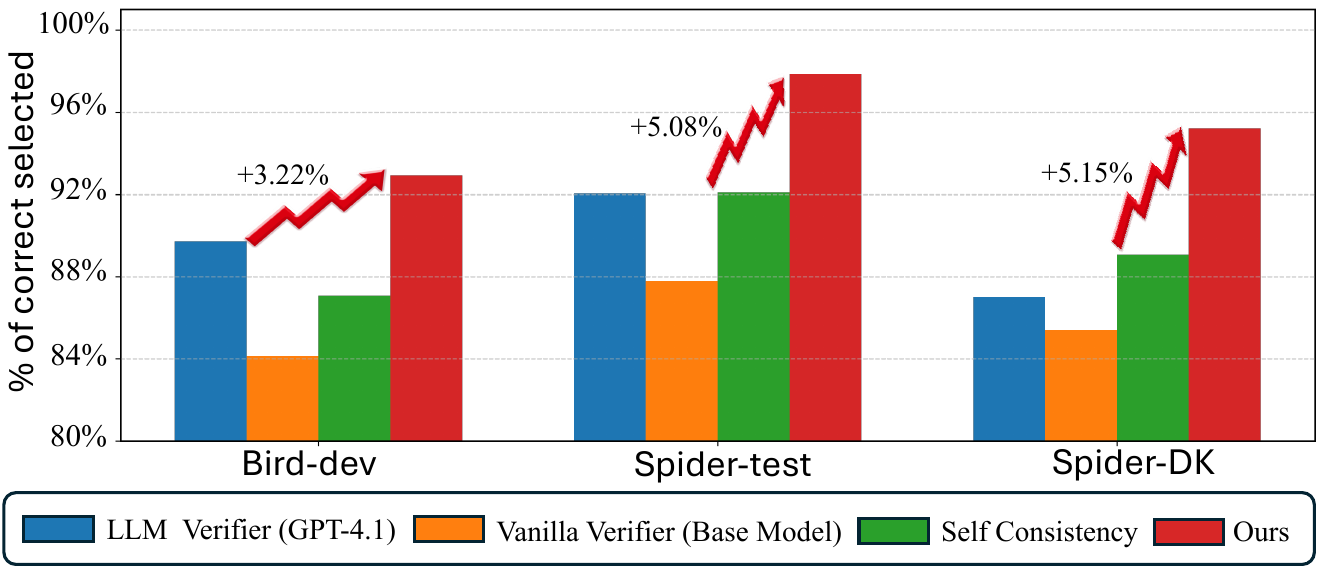}
    \caption{Comparison of different selection strategies.}
    \label{fig:verifier}
     \vspace{-0.1cm}
\end{figure}

\textbf{Selection methods analysis.}

To validate the effectiveness of our Generative Validation Agent, we compare it against several alternative selection strategies, as illustrated in Figure~\ref{fig:verifier}. While common approaches such as Self-Consistency or using a powerful LLM as a Judge (e.g., GPT-4.1) provide a reasonable baseline, their performance is suboptimal and inconsistent across the different benchmarks. In contrast, our fine-tuned Generative Validation Agent consistently outperforms all other methods by a significant margin. On the challenging Spider-test, it achieves a correct selection rate of 97.15\%, a substantial improvement over the next-best strategy's 92.09\%. Similar significant gains are observed on both BIRD-dev and Spider-DK. This consistent superiority demonstrates the stability and robustness of our specialized approach. Unlike general-purpose models or heuristic-based methods, our Validation Agent reliably identifies the most accurate reasoning trajectory, making it a critical component for achieving state-of-the-art performance. Full execution accuracy results are detailed in Appendix~\ref{Selection method comparison}.

\section{Related Work}
\textbf{LLMs for Text-to-SQL}

The rise of Large Language Models (LLMs) has brought notable progress to Text-to-SQL tasks, moving past traditional sequence-to-sequence approaches. Recent studies emphasize in-context learning, where strategies such as Chain-of-Thought (CoT) prompting are used to break down complex queries into intermediate reasoning steps~\citep{tai2023exploringchainofthoughtstyleprompting,dong2023c3zeroshottexttosqlchatgpt}. Frameworks like DIN-SQL~\citep{pourreza2023dinsqldecomposedincontextlearning} and DAIL-SQL~\citep{gao2023texttosqlempoweredlargelanguage} have systematically explored prompt engineering and multi-stage pipelines that include schema linking, generation, and refinement to boost performance. Building on these ideas, more recent studies ~\citep{wang2025macsqlmultiagentcollaborativeframework,deng2025reforce,gao2025previewxiyansqlmultigeneratorensemble,xie2025opensearchsqlenhancingtexttosqldynamic,guo2025mtsql} move toward structured, multi-step workflows that better match the complexity of real databases and diverse queries. Notably, while works like \textbf{MSc-SQL}~\citep{gorti2025msc} employ multi-sample critiquing via supervised fine-tuning, our work shifts away from fixed datasets or static prompting by introducing a dynamic, reinforcement-learning-based agentic system that optimizes interaction trajectories directly against execution feedback.

\textbf{Multi-Agent systems}

Large Language Models (LLMs) have enabled sophisticated multi-agent systems by adopting specialized roles via in-context prompting~\citep{wang_incharacter_2024,min_rethinking_2022}. Our focus is on goal-oriented problem-solving frameworks, rather than social simulations~\citep{zhang2024chainagentslargelanguage, hua_war_2024,yuan2023craft}, where tasks are divided among collaborating agents. The complexity of these collaborations has grown from simple debating~\citep{du2023improvingfactualityreasoninglanguage} to structured workflows with the use of tools, such as software development agents ChatDev~\citep{qian_chatdev_2024}, MedTutor-R1 ~\citep{he2025medtutorr1socraticpersonalizedmedical} and CollabUIAgent~\citep{he_advancing_2025}. Other notable approaches include the generic framework AutoGen~\citep{wu2023autogenenablingnextgenllm} and the dynamic cooperation in AutoAgents~\citep{chen_autoagents_2024}. Following this established paradigm, we propose a specialized pipeline for Text-to-SQL using Grounding, Generation, and Validation agents. 

\textbf{Reinforcement Learning}

Reinforcement Learning (RL) is increasingly used to enhance the complex reasoning capabilities of LLMs, especially when combined with chain-of-thought prompting ~\citep{wei2023chainofthoughtpromptingelicitsreasoning,openai2024openaio1card,su2025thinking}. This approach has proven highly effective, achieving state-of-the-art results in fields like mathematics and code generation~\citep{qin_toolllm_2023,zhao_marco-o1_2024}. Typical approaches fine-tune models with policy gradient methods such as PPO or GRPO, rewarding logical soundness or correct outcomes ~\citep{shao_deepseekmath_2024,deepseekai2025deepseekr1incentivizingreasoningcapability,huang2025adactrl,he2025medtutorr1socraticpersonalizedmedical}. While PPO is a common choice, GRPO offers advantages by being less prone to high variance and more memory-efficient, as it does not require loading an additional critic model. In parallel, interactive reasoning paradigms like ReAct~\citep{yao2022react} leverage prompting-based Think–Act–Observe loops to enable tool use and self-correction, but without explicit policy training. While Text-to-SQL requires similarly complex reasoning, explicit RL for this domain remains underexplored. Our work addresses this gap by training the Generation agent’s policy with execution-based rewards, enabling robust, stateful query generation and dynamic self-correction.

\textbf{Test-Time Scaling}

To enhance performance without the cost of retraining, many researchers have focused on inference-time techniques. Self-consistency, for instance, has become a popular method where multiple reasoning paths are sampled and the final answer is chosen by majority vote~\citep{wang_self-consistency_2023}. This concept has been further refined by verification and reranking methods, which employ an external mechanism or model to score and select the best candidate from a pool of outputs~\citep{zheng_judging_2023,gu_survey_2025}. Our approach builds on the recent innovation of Generative Verifiers~\citep{zhang_generative_2025}. Instead of a voting process or a separate classifier, our Validation Agent reframes selection as a next-token prediction problem. It assesses each potential solution trajectory by calculating the probability of the model generating a ``Yes" token, ultimately selecting the trajectory with the highest confidence score.

\section{Conclusion}
In this work, we introduced \textbf{MARS-SQL}, the first framework to leverage an \textbf{interactive RL policy for tool-augmented SQL generation under sparse execution rewards} within a multi-agent decomposition. Unlike static, single-pass approaches, our method uniquely couples this dynamic generation policy, which learns to reason and self-correct via a ReAct-style loop, with a \textbf{generative probability-based trajectory selection} mechanism tailored for SQL verification. This unified methodology enables \textbf{MARS-SQL} to establish new state-of-the-art execution accuracies on BIRD (77.84\%) and Spider (89.75\%), demonstrating exceptional cross-domain generalization by achieving top-tier performance on Spider without any in-domain training. Our ablation studies confirm the criticality of this synergistic design, suggesting that shifting from prompt engineering to interactive, policy-driven reasoning is a pivotal direction for building dependable, data-centric AI systems.

At the same time, we emphasize an important practical limitation. Although each module is 7B-scale, the deployed system contains three specialized agents and therefore has an inference-time footprint of approximately 21B parameters. This introduces a real trade-off between multi-agent specialization and single-model scaling. Our new analyses show that the observed accuracy gains are not explained by parameter count alone, but the deployment cost of the full pipeline remains higher than that of a single 7B model. We therefore view transparent reporting of total parameters, latency, token usage, and VRAM as essential for future work on efficient Text-to-SQL agents.
\newpage
\section*{Impact Statement}
This paper introduces MARS-SQL, a multi-agent framework designed to advance the field of Machine Learning by enabling more robust and self-correcting autonomous agents in complex data environments. By reframing Text-to-SQL as an interactive decision-making process grounded in reinforcement learning, our work significantly lowers the technical barriers for non-expert users to interact with large-scale, professional databases, thereby fostering the democratization of data-driven insights across diverse sectors. The specialized roles of grounding, generation, and validation agents directly address the critical societal need for more dependable AI systems, specifically by mitigating common risks such as model hallucinations and the propagation of incorrect query results in real-world applications. While this technology empowers individuals to perform complex data analysis without advanced programming knowledge, it also emphasizes a move toward transparent and verifiable AI, encouraging a deployment model that balances automation with human-in-the-loop oversight to ensure the ethical and responsible use of generated information.

\bibliography{icml_main}
\bibliographystyle{icml2026}

\newpage
\appendix
\onecolumn
\section{The Use of Large Language Models}
Large Language Models (LLMs) were utilized in a limited, assistive capacity for specific tasks in this project. For manuscript preparation, the authors supplied their own draft to an LLM, which then provided suggestions to improve grammar, enhance clarity, and ensure an academic tone. The LLM was also used to generate a list of potential titles for inspiration, though the final title was conceived and refined by the authors and not taken directly from any single output. In the implementation phase, an LLM served as a coding assistant by offering code completions and debugging support. However, all final code, experimental design, and validation were implemented and verified exclusively by the authors. It is important to emphasize that LLMs were \textbf{NOT} used for core scientific contributions, such as generating research ideas, designing experiments, or conducting the literature review. All conceptual work and experimental design originated solely with the authors.

\section{Training Details}
\label{Training details}
This section provides the detailed hyperparameters used for training our three agents. All agents were trained on NVIDIA H800 GPUs.
\subsection{Grounding Agent}
The Grounding Agent was trained using Reinforcement Learning baesd on Qwen2.5-Coder-7B-Instruct. Its primary role is to identify the correct database schema entities relevant to the user's question. The training was conducted using the Verl framework~\citep{sheng2024hybridflow}. The hyperparameters for the RL training and data generation phases are detailed in Table~\ref{tab:grounding_sql_hyperparams}.

\textbf{Training Steps and Convergence:} We trained the Grounding Agent for \textbf{600 steps} with a batch size of 64. During training, we observed clear stability and convergence patterns in the reward curves; the reward consistently increased and then plateaued, indicating that the policy was successfully optimized.

\subsection{Generation Agent}
The Generation Agent was also trained using Reinforcement Learning based on Qwen2.5-Coder-7B-Instruct, leveraging a training framework adapted from SkyRL~\citep{liu2025skyrlsql}. This agent is responsible for generating the reasoning SQL query trajectories. Its training and data generation hyperparameters are identical to those of the Grounding Agent, as shown in Table~\ref{tab:grounding_sql_hyperparams}. 

\textbf{Training Steps and Convergence:} This agent was trained for \textbf{160 steps} with a batch size of 64. Similar to the Grounding Agent, the reward curve demonstrated stable convergence within this efficient training phase.

\begin{minipage}[t!]{.48\textwidth} 
    \centering
    \captionsetup{justification=centering}
    \captionof{table}{Hyperparameters for Grounding and Generation Agent RL Training.}
    \label{tab:grounding_sql_hyperparams}
    \begin{tabular}{ll}
    \toprule
    \textbf{Parameter} & \textbf{Value} \\
    \midrule
    \multicolumn{2}{l}{\textit{Training Parameters}} \\
    \midrule
    Learning Rate & $1 \times 10^{-6}$ \\
    Batch Size & 128 \\
    \midrule
    \multicolumn{2}{l}{\textit{Trajectory Rollout Parameters}} \\
    \midrule
    Temperature & 0.6 \\
    Top-p & 0.95 \\
    \bottomrule
    \end{tabular}
\end{minipage}%
\hfill 
\begin{minipage}[t!]{.48\textwidth}
    \centering
    \captionsetup{justification=centering}
    \captionof{table}{Hyperparameters for Validation Agent Dataset Generation.}
    \label{tab:verify_data_gen_hyperparams}
    \begin{tabular}{ll}
    \toprule
    \textbf{Parameter} & \textbf{Value} \\
    \midrule
    Candidates per Question & 16 \\
    Temperature & 0.7 \\
    Top-p & 0.9 \\
    Top-k & 50 \\
    \bottomrule
    \end{tabular}
\end{minipage}

\subsection{Validation Agent}
The Validation Agent was trained via Supervised Fine-tuning (SFT) to select the best SQL query from the candidates generated by the Generation Agent. We performed a full-parameter fine-tuning of the \texttt{Qwen2.5-Coder-7B-Instruct} model~\citep{hui_qwen25-coder_2024} using the Llama Factory framework. The SFT training hyperparameters are listed in Table~\ref{tab:verify_hyperparams}, and the parameters for generating its training dataset are in Table~\ref{tab:verify_data_gen_hyperparams}.

\begin{table}[h!]
\centering
\caption{Hyperparameters for Validation Agent SFT.}
\label{tab:verify_hyperparams}
\begin{tabular}{ll}
\toprule
\textbf{Parameter} & \textbf{Value} \\
\midrule
Base Model & \texttt{Qwen2.5-Coder-7B-Instruct} \\
Epochs & 3 \\
Learning Rate Scheduler & Cosine \\
Initial Learning Rate & $1 \times 10^{-5}$ \\
Effective Batch Size & 4 \\
\quad \textit{Per-device Batch Size} & \textit{1} \\
\quad \textit{Gradient Accumulation} & \textit{2 steps} \\
Precision & \texttt{bf16} \\
Optimization & DeepSpeed ZeRO Stage 3 \\
\bottomrule
\end{tabular}
\end{table}

\section{Dataset}
\subsection{Training Dataset}
\label{Training Dataset}
Our training data is derived from the Bird benchmark, which comprises 9,428 question-SQL pairs. To ensure high quality, we first filtered this dataset by removing samples flagged as incorrect ~\citep{pourreza_reasoning-sql_2025, li_codes_2024} by both Gemini-2.5-pro and GPT-4o, resulting in a clean set of 8,036 training examples. From this set, we constructed the fine-tuning data for the grounding task. For each of the 8,036 question-database pairs, we generated a distinct training instance for every table within that database. This process resulted in a large-scale dataset of 90,102 individual data points. For each point, the ground truth—whether a table is relevant and which of its columns are used—was programmatically extracted from the gold SQL query using the SQLGlot parser. 

We constructed a specialized dataset for training the Verifier via Supervised Fine-Tuning (SFT). First, for each question in our filtered BIRD training set, we used both our fine-tuned Generator agent and the initial base model to perform inference, generating a diverse pool of 16 candidate trajectories per question. This ensures the Validation Agent is exposed to a wide range of reasoning paths, both correct and flawed. From this pool, we curated a preference dataset by selecting one positive example (a trajectory leading to a correct execution result) and one negative example (a trajectory leading to an incorrect result) for each question. We mix the order of correct and incorrect trajectories in each pair at random to prevent order bias during training. Since the number of cases containing both correct and incorrect trajectories is limited, some questions yield only flawed trajectories. In such cases, we add the ground truth SQL query in the prompt as a suggestion to help the model generate proper trajectories. We applied best-of-N and worst-of-N~\citep{gui_bonbon_2024} strategies to select both positive and negative examples. This process yielded a final dataset of approximately 16,000 training instances. Each instance is a triplet containing the user's question, the full interaction trajectory (including all [Think], [SQL], and [Observation] steps), and the final execution result.

\subsection{Evaluation Dataset}
\label{Evaluation Dataset}
BIRD is a large-scale, realistic benchmark designed to evaluate modern Text-to-SQL systems. It features complex databases (33.4 GB across 95 databases), questions from 37 professional domains, and imperfect real-world data values requiring robust handling. BIRD uniquely emphasizes the generation of both correct and efficient SQL queries, making it an ideal testbed for our framework. Our primary evaluations are performed on its development set, which contains 1,534 examples.

Spider 1.0 is a comprehensive, cross-domain benchmark containing 10,181 questions and 5,693 unique complex SQL queries across 200 multi-table databases. It serves as a standard for evaluating cross-domain Text-to-SQL performance. For our evaluation, we use the official test set, which includes 2,147 examples.

Spider-DK, an extension of Spider, is designed specifically to test a model's ability to handle queries requiring implicit domain knowledge. It comprises samples from the Spider development set that were manually modified to depend on real-world information for correct interpretation. This benchmark simulates scenarios where user queries rely on specific domain context. We evaluate our model on the Spider-DK test set, which contains 535 examples.

SParC~\citep{yu2019sparccrossdomainsemanticparsing} is a conversational and cross-domain Text-to-SQL benchmark in which each user question depends on the preceding interaction context. We use it as an additional robustness check because it stresses both cross-domain transfer and multi-turn understanding.

\section{Additional Out-of-Domain Robustness}
\label{sec:additional_ood}

To complement the main-paper results on Spider-test and Spider-DK, we further evaluate MARS-SQL on SParC. Table~\ref{tab:additional_ood} shows that our model remains strong across these out-of-domain settings, including a particularly large gain on SParC, a benchmark that stresses contextual and interaction-heavy reasoning.

\begin{table}[h]
    \centering
    \caption{Additional out-of-domain robustness results. All numbers are execution accuracy (\%).}
    \label{tab:additional_ood}
    \begin{tabular}{lccc}
        \toprule
        \textbf{Model} & \textbf{Spider-test} & \textbf{Spider-DK} & \textbf{SParC} \\
        \midrule
        GPT-O3-mini & 78.82 & 71.77 & 67.0 \\
        GPT-5 & 77.10 & 66.92 & 61.8 \\
        Qwen2.5-7B & 75.87 & 61.31 & 64.1 \\
        \textbf{MARS-SQL (Ours)} & \textbf{89.75} & \textbf{78.13} & \textbf{85.78} \\
        \bottomrule
    \end{tabular}
\end{table}

\section{Training Efficiency and Resource Analysis}
\label{sec:training_cost}

To address concerns regarding the computational resources required for our multi-agent framework, we provide a detailed breakdown of the training time and a comparative analysis of data efficiency. All experiments were conducted on a node equipped with 4 $\times$ NVIDIA H800 GPUs.

\subsection{Computational Cost Breakdown}
Contrary to the perception that training multiple agents is prohibitively resource-intensive, our framework is designed for rapid convergence. As detailed in Table~\ref{tab:training_breakdown}, the entire specialized training pipeline—including the SFT for the Validation Agent and GRPO-based Reinforcement Learning for both the Grounding and Generation Agents—completes in approximately \textbf{13 hours}. This represents a modest one-time computational cost, especially considering the significant performance gains achieved.

\begin{table}[h]
    \centering
    \caption{ One-Time Training Cost breakdown on 4 $\times$ NVIDIA H800 GPUs.}
    \label{tab:training_breakdown}
    \begin{tabular}{l c c c c}
        \toprule
        \textbf{Agent} & \textbf{Method} & \textbf{Training Steps} & \textbf{Batch Size} & \textbf{Est. Training Time} \\
        \midrule
        Validation Agent & SFT & $\sim$10k & 4 & 1 h \\
        Grounding Agent & GRPO & 600 & 64 & 4 h \\
        Generation Agent & GRPO & 160 & 64 & 8 h \\
        \midrule
        \textbf{Total} & & & & \textbf{$\sim$13 h} \\
        \bottomrule
    \end{tabular}
\end{table}

\subsection{Data Efficiency and Comparative Analysis}
The efficiency of MARS-SQL stems from its ability to learn diverse reasoning and self-correction behaviors through interaction and self-play, rather than relying on massive-scale supervised datasets. 

Table~\ref{tab:efficiency_comparison} compares our framework against standard single-agent SFT approaches. While standard SFT on the BIRD training set (12k examples) takes only ~2 hours, it yields a significantly lower execution accuracy (EX) of 61.08\%. Scaling up SFT, as seen in methods like OminiSQL (utilizing 2.5M examples), requires approximately 20 days of training yet only reaches 64.50\% EX.

In contrast, MARS-SQL achieves a state-of-the-art EX of \textbf{77.84\%} using only \textbf{35k} LLM-labeled examples and 13 hours of training. To match this performance level using a single-agent SFT-only paradigm, we conservatively estimate—based on scaling laws—that it would require approximately \textbf{15 million} synthetic examples and \textbf{3--4 months} of training time on the same hardware. Thus, our multi-agent RL framework offers orders of magnitude better data and compute efficiency.

\begin{table}[h]
    \centering
    \caption{ Cost and efficiency analysis compared with single-agent SFT baselines on Bird-Dev.}
    \label{tab:efficiency_comparison}
    \begin{tabular}{l l c r c}
        \toprule
        \textbf{Method} & \textbf{Annotation Source} & \textbf{Data Size} & \textbf{Training Time (wall)} & \textbf{Dev EX (\%)} \\
        \midrule
        Original (Baseline) & --- & --- & --- & 54.56 \\
        SFT on BIRD-train & Human & 12,000 & $\sim$2 h & 61.08 \\
        Large SFT (e.g., OminiSQL) & LLM + Human & 2,500,000 & $\sim$20 days & 64.50 \\
        \midrule
        \textbf{MARS-SQL (Ours)} & LLM & \textbf{35,000} & \textbf{$\sim$13 h} & \textbf{77.84} \\
        \bottomrule
    \end{tabular}
\end{table}

\section{Inference Efficiency and Practicality Analysis}
\label{sec:appendix_efficiency}
In this section, we provide a comprehensive analysis of the efficiency and practicality of the MARS-SQL framework. Our goal is not to argue that all systems should be compared as if they had identical product characteristics. Instead, we characterize the end-to-end cost of MARS-SQL and analyze whether its additional test-time compute translates into meaningful accuracy gains compared with both Text-to-SQL baselines and general reasoning models. Since \textbf{MARS-SQL} is deployed as three specialized 7B agents, we additionally report the total inference-time parameter footprint ($\sim$21B) and the practical VRAM implications of sequential versus fully online deployment. Our analysis covers five aspects: (1) baseline latency and token cost, (2) memory footprint, (3) performance under normalized time budgets, (4) adjustable cost--accuracy trade-offs, and (5) potential system-level optimizations.

\subsection{Baseline Time and Token Cost Analysis}
We first present the latency breakdown for our standard SOTA-performing configuration ($N_g=8$ trajectories, $N_v=8$ validation samples). As shown in Table~\ref{tab:latency_breakdown}, the average end-to-end latency is 22.12 seconds per query to achieve 77.84\% accuracy.

\begin{table}[h]
    \centering
    \small
    \caption{ Average End-to-End Latency Analysis of MARS-SQL on the BIRD dev set (Hardware: 1x A6000, num\_cpus=32). Times represent the average latency to generate one SQL query.}
    \begin{tabular}{l c l}
    \toprule
    \textbf{ Stage} & \textbf{ Avg. Time (s)} & \textbf{ Description} \\
    \midrule
     1. Grounding Agent &  0.78s &  Aggregated calls (1 per table) \\
     2. Generation Agent &  18.77s &  Generating $N_g = 8$ trajectories \\
     3. Validation Agent &  2.58s &  Validating $N_g = 8$ trajectories ($N_v = 8$ samples each) \\
    \textit{ Ref: SQL Exec Time} & \textit{ (2.37s)} & \textit{ Avg. time to execute the ground truth SQL} \\
    \midrule
    \textbf{ Total (End-to-End)} & \textbf{ 22.12s} & \textbf{ Sum of all stages} \\
    \bottomrule
    \end{tabular}
    \label{tab:latency_breakdown}
\end{table}
The token consumption is analyzed in Table~\ref{tab:token_cost}. The Generation Agent, utilizing a multi-turn ``Think-Act-Observe'' loop, accounts for the majority of the token usage.

\begin{table}[h]
    \centering
    \small
    \caption{ Average Token Cost Analysis per Query.}
    \begin{tabular}{l c l}
    \toprule
    \textbf{ Stage} & \textbf{ Avg. Tokens} & \textbf{ Description} \\
    \midrule
     1. Grounding Agent &  875 &  Prompt + Schema + Question + Output \\
     2. Generation Agent &  9,200 &  $N_g=8 \times$ (Prompt + Schema + Question + Traj.) \\
     3. Validation Agent &  3,250 &  $N_g=8 \times N_v=8 \times$ (Prompt + Trajectory) \\
    \midrule
    \textbf{ Total (Avg.)} & \textbf{ 13,325} & \textbf{ Sum of all components} \\
    \bottomrule
    \end{tabular}
    \label{tab:token_cost}
\end{table}

\subsection{Memory Footprint and Deployment Implications}

The latency and token analyses above do not fully characterize deployment cost, because \textbf{MARS-SQL} is a three-agent system. We therefore report the memory footprint of the full pipeline under a HuggingFace-based sequential deployment that avoids additional engine-level pre-allocation. In this setup, the three agents are executed sequentially (Grounder $\rightarrow$ Generator $\rightarrow$ Validation), so the peak VRAM is close to that of a single active 7B agent rather than the sum of all three models loaded simultaneously.

\begin{table}[h]
    \centering
    \small
    \caption{VRAM footprint of MARS-SQL under different deployment assumptions.}
    \label{tab:vram_footprint}
    \begin{tabular}{lcc}
    \toprule
    \textbf{Deployment Setting} & \textbf{VRAM} & \textbf{Note} \\
    \midrule
    Single 7B checkpoint (BF16 weights on disk) & 15.22 GB & Static checkpoint size \\
    Sequential inference peak (2K context) & $\approx$18.5 GB & One active agent at a time \\
    Estimated full online deployment (3 agents resident) & $\approx$46 GB & Sum of weights + runtime buffers \\
    A6000 + SGLang (observed) & $\approx$41 GB & Includes engine pre-allocation \\
    A6000 + vLLM (observed) & $\approx$47 GB & Includes engine pre-allocation \\
    \bottomrule
    \end{tabular}
\end{table}

This distinction matters for interpreting the ``7B-class'' description. Each module is indeed built from a 7B model, but the total inference-time system footprint is approximately 21B parameters. At the same time, peak VRAM in our default sequential execution is much closer to a single 7B deployment than to loading all three modules simultaneously. In other words, parameter count and peak memory are related but not identical metrics in our setting.

We also emphasize that the higher VRAM observed with SGLang or vLLM on A6000 GPUs reflects system-level optimization choices such as memory reservation and kernel/workspace pre-allocation for throughput, rather than the intrinsic minimum memory required by the model weights themselves.

\subsection{Accuracy--Latency Trade-offs}

\begin{figure}[h]
    \centering
    \includegraphics[width=0.7\linewidth]{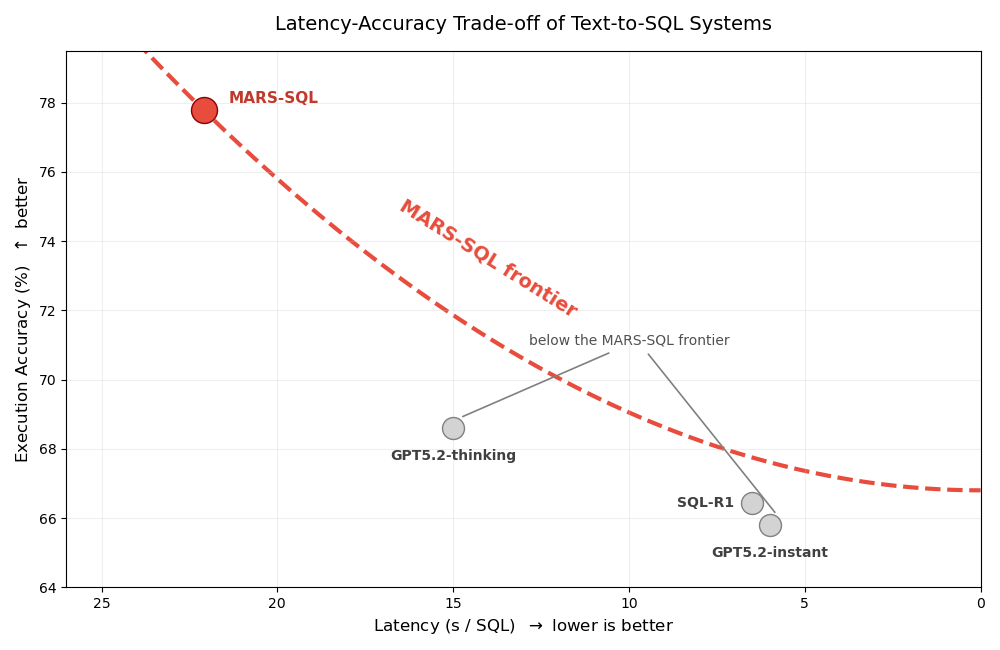}
    \caption{Accuracy--latency Pareto plot using the methods shown in Table~\ref{tab:latency_tradeoff_closed}. MARS-SQL lies on the high-accuracy frontier and extends the same trend observed in reasoning models: more test-time compute can be worthwhile when it buys substantially better execution accuracy.}
    \label{fig:pareto}
\end{figure}

We next analyze whether the additional latency of MARS-SQL translates into meaningful performance gains. Our GPT-5.2 comparison is \emph{not} intended as a direct system-level comparison with MARS-SQL. Instead, it is used to illustrate the efficiency of test-time compute trade-offs within a similar model family: additional ``thinking'' time is valuable when it yields sufficiently large accuracy gains. Table~\ref{tab:latency_tradeoff_closed} summarizes this comparison under a unified local profiling setup, while Figure~\ref{fig:pareto} visualizes the same trend as an accuracy--latency frontier. From this perspective, MARS-SQL follows the same principle but delivers a stronger return on added latency for Text-to-SQL. To further address the concern that a strong single-agent baseline might catch up given a comparable inference budget, we also evaluate a high-budget SQL-R1 setting with 64 sampled outputs and majority voting.

\begin{table}[h]
    \centering
    \small
    \caption{Accuracy--latency trade-off under a unified local profiling setup. ``Total Time'' includes generation and SQL execution.}
    \label{tab:latency_tradeoff_closed}
    \begin{tabular}{lcc}
    \toprule
    \textbf{Method} & \textbf{Total Time (s)} & \textbf{EX Acc. (\%)} \\
    \midrule
    GPT-5.2-instant & 6.0 & 65.8 \\
    GPT-5.2-thinking & 15.0 & 68.6 \\
    SQL-R1 & 6.5 & 66.43 \\
    SQL-R1 + 64 Samples + Voting & 33.8 & 67.34 \\
    \textbf{MARS-SQL (Ours)} & \textbf{22.1} & \textbf{77.8} \\
    \bottomrule
    \end{tabular}
\end{table}

This interpretation is further supported by the incremental accuracy-per-second return of extra test-time compute. GPT-5.2-instant $\rightarrow$ GPT-5.2-thinking yields approximately $+0.31$ acc/sec, while SQL-R1 $\rightarrow$ MARS-SQL yields approximately $+0.73$ acc/sec. Importantly, even when SQL-R1 is given a substantially larger test-time budget than MARS-SQL (33.8s vs.\ 22.1s), its final majority-voted accuracy remains only 67.34\%. In other words, MARS-SQL does not simply spend more time; it converts additional latency into larger execution-accuracy gains on challenging Text-to-SQL reasoning, which is also visible in Figure~\ref{fig:pareto}.

To further separate this observation from any product-level benefits unique to proprietary systems, Table~\ref{tab:latency_tradeoff_open} reports the same style of comparison on open-weight reasoning models. Across Qwen3-30B-A3B, gpt-oss-20B, and GPT-5.2, enabling more deliberative inference consistently trades latency for higher accuracy. MARS-SQL occupies the high-accuracy end of the same trend, and also shows the highest accuracy gain per added second among the comparisons reported here.

\begin{table}[h]
    \centering
    \small
    \caption{Open-weight and reasoning-model comparisons under the same profiling setup.}
    \label{tab:latency_tradeoff_open}
    \begin{tabular}{lccc}
    \toprule
    \textbf{Method} & \textbf{Time (s)} & \textbf{Acc. (\%)} & \textbf{Note} \\
    \midrule
    Qwen3-30B-A3B (No Thinking) & 1.3 & 48.11 & --- \\
    Qwen3-30B-A3B (With Thinking) & 15.8 & 54.69 & +6.58\% ACC for +14.5s \\
    gpt-oss-20B (No Thinking) & 3.6 & 52.74 & --- \\
    gpt-oss-20B (With Thinking) & 12.4 & 55.80 & +3.06\% ACC for +8.8s \\
    GPT-5.2-instant (No Thinking) & 6.0 & 65.8 & --- \\
    GPT-5.2-thinking (With Thinking) & 15.0 & 68.6 & +2.8\% ACC for +9.0s \\
    SQL-R1 & 6.5 & 66.43 & --- \\
    SQL-R1 + 64 Samples + Voting & 33.8 & 67.34 & +0.91\% ACC for +27.3s \\
    \textbf{MARS-SQL (Ours)} & \textbf{22.1} & \textbf{77.8} & \textbf{+11.4\% ACC for +15.6s vs. SQL-R1} \\
    \bottomrule
    \end{tabular}
\end{table}

\subsection{Performance Comparison Under Normalized Time Budget}

The accuracy--latency curves above compare several operating regimes, but one reviewer-facing concern is even more direct: can a strong single-agent system match MARS-SQL if it is granted the same wall-clock budget? To answer this, we additionally compare against self-consistency baselines under an equal time budget of approximately 22 seconds, matching the standard MARS-SQL configuration ($N_g=8, N_v=8$). The SFT baseline uses the budget to generate 16 independent samples, while the closed-source baseline uses the budget to make four independent API calls; both select the final answer by self-consistency.

\begin{table}[h]
    \centering
    \small
    \caption{Accuracy comparison under a normalized time budget ($\approx$22s).}
    \begin{tabular}{llcc}
    \toprule
    \textbf{Method} & \textbf{Configuration} & \textbf{Avg. Latency} & \textbf{EX Acc. (\%)} \\
    \midrule
    \texttt{Qwen-SFT} (Self-Consistency) & SFT + 16 samples & $\approx$22.0s & 64.2 \\
    \texttt{GPT-5} (Self-Consistency) & 4 API calls & $\approx$22.0s & 69.3 \\
    \textbf{MARS-SQL (Ours)} & \textbf{Multi-agent RL} & \textbf{22.12s} & \textbf{77.84} \\
    \bottomrule
    \end{tabular}
    \label{tab:time_normalized}
\end{table}

As shown in Table~\ref{tab:time_normalized}, these baselines fail to match the performance of MARS-SQL even when granted an equivalent time budget. This supports the view that the gain comes from the structured multi-agent reasoning workflow and trained trajectory validation, rather than merely from spending more inference time.

\subsection{Adjustable Cost--Accuracy Trade-offs}

The latency reported in Table~\ref{tab:latency_breakdown} represents a performance-oriented configuration. MARS-SQL allows for flexible deployment by adjusting the number of generation trajectories ($N_g$) and validation samples ($N_v$). Table~\ref{tab:cost_tradeoff} illustrates these trade-offs.

\begin{table}[h]
    \centering
    \caption{ Tunable Cost-Accuracy Curve for MARS-SQL.}
    \begin{tabular}{lcccc}
    \toprule
    \textbf{ Mode} & \textbf{ Params ($N_g, N_v$)} & \textbf{ Latency} & \textbf{ Acc. (\%)} & \textbf{ Characteristic} \\
    \midrule
     Fast &  (1, 1) &  3.1s &  68.71\% &  High speed, outperforms SFT \\
     Balanced &  (4, 4) &  11.5s &  74.90\% &  Balanced cost-benefit \\
    \textbf{ SOTA (Ours)} & \textbf{ (8, 8)} & \textbf{ 22.1s} & \textbf{ 77.84\%} & \textbf{ Maximum accuracy} \\
     Over-Sampling &  (16, 8) &  42.8s &  77.84\% &  Diminishing returns \\
    \bottomrule
    \end{tabular}
    \label{tab:cost_tradeoff}
\end{table}

Users can select a ``Fast'' setting to achieve a $\approx$3-second response that still surpasses the greedy SFT baseline, or invest more computational resources for maximum performance.

We do not claim that MARS-SQL is universally cheaper than every single larger model. Rather, our claim is that its accuracy gains are not explained by parameter scaling alone, and that its deployment trade-offs can be made explicit through latency, token, and VRAM reporting.

\subsection{System-Level Optimization}

The latency metrics presented above assume a sequential, single-query execution, serving as a conservative upper bound. In practical multi-user deployments, \textbf{MARS-SQL} can achieve higher throughput through system-level optimizations:

\begin{enumerate}
    \item \textbf{Pipeline Parallelism:} The Grounder, Generator, and Validator agents can process different queries in parallel, creating a pipeline for incoming requests.
    \item \textbf{Batched Validation:} The $N_g \times N_v$ validation calls are embarrassingly parallel and can be fused into batched requests to reduce amortized costs.
\end{enumerate}

\begin{table}[h]
    \centering
    \small
    \caption{ Sequential vs. System-Optimized Deployment (Conceptual Comparison).}
    \begin{tabular}{p{3.5cm} p{4.5cm} c c}
    \toprule
    \textbf{ Deployment} & \textbf{ Execution Pattern} & \textbf{ Est. Latency} & \textbf{ Est. Throughput} \\
    \midrule
    \textbf{ Sequential} \newline  (No optimization) &  Grounder $\to$ Generator $\to$ Validator (End-to-End) &  $\approx 22.1$s &  $\approx 2.7$ queries/min \\
    \midrule
    \textbf{ System-Optimized} \newline  (Pipeline + Batch) &  Pipelined stages; batched validation &  $\approx 12\text{--}15$s &  $\approx 4\text{--}5$ queries/min \\
    \bottomrule
    \end{tabular}
    \label{tab:system_opt}
\end{table}

Table~\ref{tab:system_opt} estimates that with these optimizations, the effective per-query latency can be reduced by approximately 40--60\%, significantly improving throughput on a single GPU node.

\section{Architecture versus Parameter Scaling}
\label{sec:param_scaling_appendix}
A critical consideration when evaluating multi-agent frameworks is whether the observed performance improvements are intrinsically tied to the novel architectural design, or merely a byproduct of increased parameter capacity at inference time. To decouple our methodological contributions from pure scaling effects, we conduct three complementary analyses. First, we benchmark \textbf{MARS-SQL} against a parameter-matched ($\sim$21B) ensemble constructed from highly competitive 7B Text-to-SQL models (Table~\ref{tab:ensemble_21b}). Second, we investigate the scaling behavior of the core framework by expanding the Generation Agent from 7B to 14B parameters, maintaining the identical Reinforcement Learning (RL) training recipe (Table~\ref{tab:scaling_vs_specialization}). Third, to directly address the possibility that a single larger model could absorb all three roles through increased capacity, we train single-model multi-task baselines that use one shared model to perform grounding, generation, and validation (Table~\ref{tab:single_model_multitask}).
\begin{table}[h]
\centering
\small
\caption{Comparison against a strong $\sim$21B ensemble on BIRD-dev. The ensemble combines Arctic-text2sql-7B, SQL-R1, and OmniSQL-7B, generates eight queries per model, and selects the final answer by majority voting.}
\label{tab:ensemble_21b}
\begin{tabular}{lcc}
\toprule
\textbf{System Configuration} & \textbf{Total Parameters} & \textbf{EX Acc. (\%)} \\
\midrule
Arctic-text2sql-7B (Single) & 7B & 67.73 \\
SQL-R1 (Single) & 7B & 66.03 \\
OmniSQL-7B (Single) & 7B & 65.31 \\
\textbf{SOTA 7B Ensemble (Voting)} & \textbf{$\sim$21B} & \textbf{72.10} \\
\textbf{MARS-SQL (Ours)} & \textbf{$\sim$21B} & \textbf{77.84} \\
\bottomrule
\end{tabular}
\end{table}
\begin{table}[h]
\centering
\small
\caption{Scaling the generator versus using multi-agent specialization on BIRD-dev.}
\label{tab:scaling_vs_specialization}
\begin{tabular}{lccc}
\toprule
\textbf{System Configuration} & \textbf{Core Model} & \textbf{EX Acc. (\%)} & \textbf{Gain over 7B Base} \\
\midrule
Generator Only (Base) & 7B & 66.37 & -- \\
Generator Only (Scaled) & 14B & 67.92 & +1.55 \\
\textbf{MARS-SQL (Ours)} & \textbf{3 $\times$ 7B} & \textbf{77.84} & \textbf{+11.47} \\
MARS-SQL (Scaled Gen) & 14B + 2 $\times$ 7B & 78.62 & +12.25 \\
\bottomrule
\end{tabular}
\end{table}
\begin{table}[H]
\centering
\small
\caption{Single-model multi-task baselines on BIRD-dev. A single shared model is SFT-trained to perform all three roles: grounding, generation, and validation.}
\label{tab:single_model_multitask}
\begin{tabular}{lcc}
\toprule
\textbf{System Configuration} & \textbf{Model Size} & \textbf{EX Acc. (\%)} \\
\midrule
Single-model multi-task (SFT) & 14B & 55.67 \\
Single-model multi-task (SFT) & 32B & 57.82 \\
\textbf{MARS-SQL (Ours)} & \textbf{3 $\times$ 7B} & \textbf{77.84} \\
\bottomrule
\end{tabular}
\end{table}
These results support three conclusions. First, as shown in Table~\ref{tab:ensemble_21b}, simply stacking three strong 7B Text-to-SQL models into a $\sim$21B ensemble reaches 72.10\%, which is still 5.74 points below \textbf{MARS-SQL}. Second, Table~\ref{tab:scaling_vs_specialization} shows that scaling the generator from 7B to 14B improves the single-agent baseline by only 1.55 points, whereas introducing the specialized multi-agent workflow improves it by 11.47 points. Third, Table~\ref{tab:single_model_multitask} demonstrates that a single shared model trained to absorb all three roles performs substantially worse even at larger scales: the 14B and 32B multi-task baselines reach only 55.67\% and 57.82\% EX, respectively. Together, these findings indicate that the dominant source of improvement is the role-specialized interactive workflow, not raw parameter count alone.

\section{Multi-turn Generation}
\label{Multi-turn Generation}
\textbf{Evolution of Interaction Turns: }
To understand the impact of RL training on the agent's reasoning efficiency, we analyzed the evolution of rollout lengths during the training process. In our setting, each ``Think--Act--Observe'' cycle corresponds to one database interaction turn, making the average number of interaction turns a proxy for rollout length.

\textbf{Grounding Agent (Single-turn):} The rollout length (token count) exhibited a mild U-shaped pattern. Initially, the output became more concise, followed by a slight lengthening to include only essential schema information. This reflects a refinement of the policy towards precise schema selection rather than reasoning from scratch.

\textbf{Generation Agent (Multi-turn):} A distinct trend was observed where the average number of interaction turns consistently decreased and stabilized at a lower level. This indicates that the agent learned to solve problems more directly and recognized when to terminate the search efficiently. This efficiency gain is quantitatively supported by the evaluation on the BIRD-dev set (with a maximum of 5 turns), as shown in Table~\ref{tab:interaction_turns}. The RL-trained agent significantly reduces the average turns across all difficulty levels compared to the base model.

\begin{table}[h]
\centering
\caption{Comparison of Average Interaction Turns on BIRD-dev (Max Turns = 5) before and after RL training.}
\label{tab:interaction_turns}

\begin{tabular}{l c c c}
\toprule
\textbf{Model} & \textbf{Avg. Turns (Challenging)} & \textbf{Avg. Turns (Moderate)} & \textbf{Avg. Turns (Simple)} \\
\midrule
Before RL (Base Model) & 2.90 & 2.67 & 2.27 \\
\textbf{After RL (Generation Agent)} & \textbf{1.82} & \textbf{1.71} & \textbf{1.45} \\
\bottomrule
\end{tabular}
\end{table}
\begin{table}[h]
\centering
\caption{Distribution of Interaction Turns Used by the Generation Agent on BIRD-dev.}
\label{tab:turn_distribution}
\begin{tabular}{c c}
\toprule
\textbf{\# of Interaction Turns} & \textbf{\# of Examples} \\
\midrule
1 & 1,116 \\
2 & 174 \\
3 & 105 \\
4 & 88 \\
5+ & 51 \\
\bottomrule
\end{tabular}
\end{table}
Furthermore, analyzing the specific distribution of turns reveals that the agent learns an adaptive and non-wasteful strategy. As presented in Table~\ref{tab:turn_distribution}, while the agent retains the capacity to use multiple turns for complex reasoning, it solves the vast majority of problems (1,116 cases) in a single interaction. Crucially, for the "long-tail" of more difficult queries, the agent robustly applies deeper reasoning, utilizing up to 5 or more turns to arrive at the correct solution. This distribution confirms that the agent is not bound by arbitrary limits but instead dynamically decides the necessary reasoning depth for each specific query.

\begin{figure}[h]
    \centering
    \caption{Example of interactive process in the generation.}
    \includegraphics[width=\textwidth]{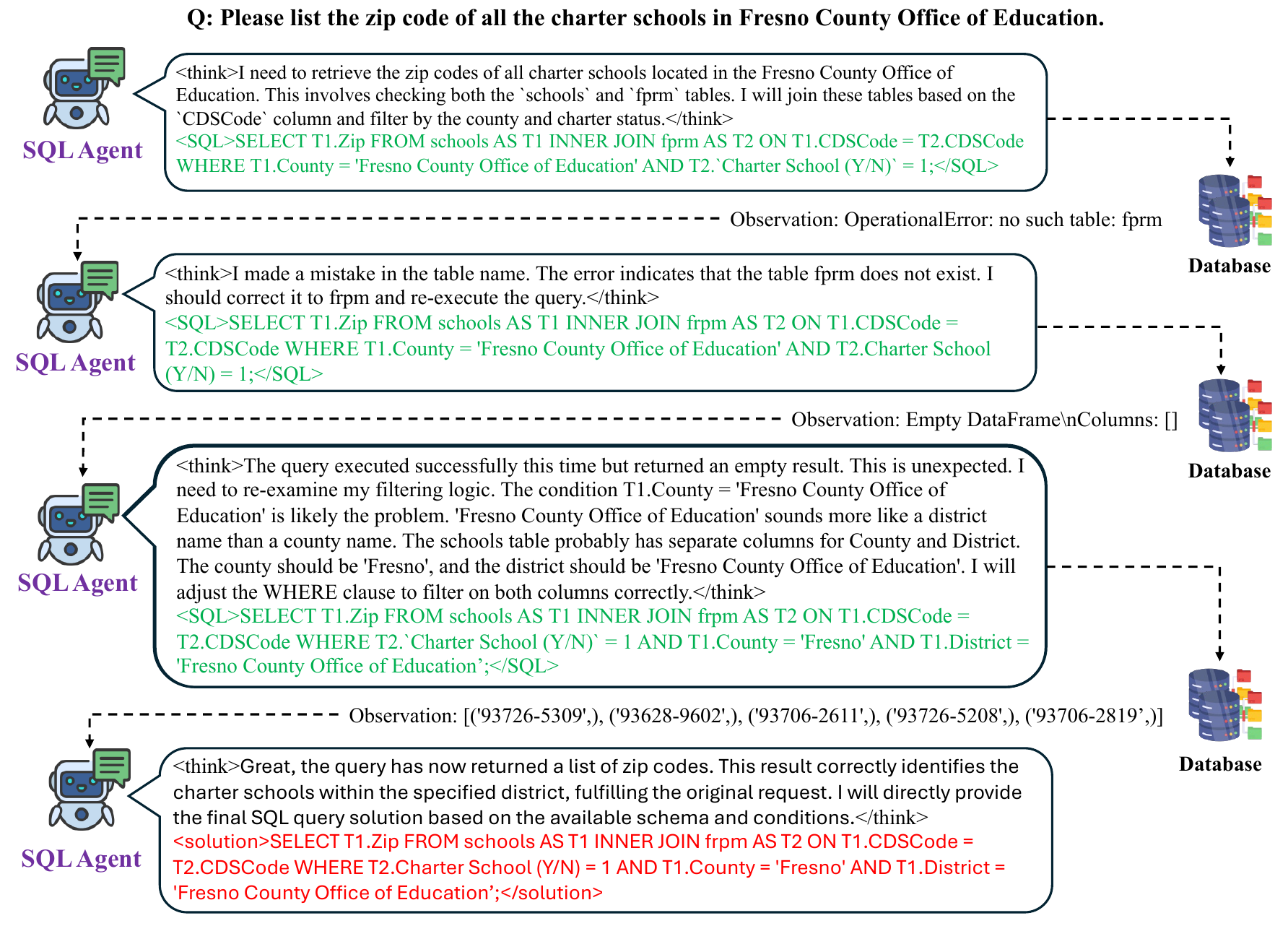}
    \label{fig:Interatctive_process}
\end{figure}

Figure~\ref{fig:Interatctive_process} provides a concrete example of the agent's interactive and self-correcting workflow. The agent initially generates a query with a typographical error in a table name `fprm'. Upon receiving an `OperationalError' from the database, it correctly identifies the mistake in its next thought process, corrects the table name to `frpm', and re-executes the query. The second attempt returns an empty result, prompting the agent to re-evaluate its filtering logic. It deduces that it incorrectly used a district name as a county name and refines the `WHERE' clause accordingly. This final attempt successfully retrieves the correct data, leading the agent to output the validated query as its final solution.

\section{Qualitative Analysis: Reasoning Beyond Valid Execution}
\label{app:reasoning_scenarios}

A core feature of our Generation Agent is its learned policy to evaluate the \textit{semantic content} and \textit{plausibility} of execution results, rather than merely checking for syntax errors. This allows the agent to continue reasoning even when the database returns a non-empty, non-error result. Below, we present two real-world case studies from the BIRD-dev set illustrating this capability.

\subsection{Scenario 1: Complex Problem Decomposition}
\label{app:scenario_decomposition}

\interlinepenalty=5000 

In complex queries, a successful execution often represents only an intermediate step in a larger plan. The agent effectively decomposes the problem, using the observation from the first step to inform the subsequent logic. 

\par\medskip
\noindent\textbf{Question:} Consider the average difference between K-12 enrollment and 15-17 enrollment of schools that are locally funded, list the names and DOC type of schools which has a difference above this average. 

\par\bigskip
\begin{minipage}{\linewidth}
\noindent\textbf{Turn 1 (Think):} To solve this problem, I need to follow these steps: 1) Calculate the average difference between K-12 enrollment and 15-17 enrollment for all schools. 2) Filter the schools based on the funding type. 3) Calculate the difference for filtered schools. 4) Identify schools above the average. 

\smallskip
\noindent\textbf{Turn 1 (SQL):}
\begin{lstlisting}[language=SQL, basicstyle=\ttfamily\small, breaklines=true, frame=single, xleftmargin=1.0em]
SELECT AVG("Enrollment (K-12)" - "Enrollment (Ages 5-17)") FROM frpm;
\end{lstlisting}

\smallskip
\noindent\textbf{Turn 1 (Observation):} \texttt{16.261833} 
\end{minipage}

\par\bigskip
\begin{minipage}{\linewidth}
\noindent\textbf{Turn 2 (Think):} \textit{(The agent recognizes this is an intermediate value, not the final answer.)} The average difference is approximately 16.26. Now, I will filter the schools based on the funding type (locally funded) and calculate the difference for these schools against this average. 

\smallskip
\noindent\textbf{Turn 2 (SQL):}
\begin{lstlisting}[language=SQL, basicstyle=\ttfamily\small, breaklines=true, frame=single, xleftmargin=1.0em]
SELECT T2.School, T2.DOC FROM frpm AS T1 JOIN schools AS T2 ON T1.CDSCode = T2.CDSCode WHERE T2.FundingType = 'Locally funded' AND (T1."Enrollment (K-12)" - T1."Enrollment (Ages 5-17)") > 16.26;
\end{lstlisting}

\smallskip
\noindent\textbf{Turn 2 (Observation):} \texttt{[(`Circle of Independent Learning', `54'), (`Learning Community Charter', `00')]} 
\end{minipage}

\par\bigskip
\begin{minipage}{\linewidth}
\noindent\textbf{Turn 3 (Think):} The query executed successfully and returned non-empty results. I will now combine the logic into a single nested query for the final answer. 

\smallskip
\noindent\textbf{Turn 3 (Final Solution):}
\begin{lstlisting}[language=SQL, basicstyle=\ttfamily\small, breaklines=true, frame=single, xleftmargin=1.0em]
SELECT T2.School, T2.DOC FROM frpm AS T1 INNER JOIN schools AS T2 ON T1.CDSCode = T2.CDSCode WHERE T2.FundingType = 'Locally funded' AND (T1."Enrollment (K-12)" - T1."Enrollment (Ages 5-17)") > (SELECT AVG(T3."Enrollment (K-12)" - T3."Enrollment (Ages 5-17)") FROM frpm AS T3 INNER JOIN schools AS T4 ON T3.CDSCode = T4.CDSCode WHERE T4.FundingType = 'Locally funded')
\end{lstlisting}
\end{minipage}
\subsection{Scenario 2: Data Plausibility and Sanity Checking}
\label{app:scenario_sanity_check}

\interlinepenalty=5000 

The agent also learns to perform ``sanity checks.'' If a syntactically correct query returns implausible data (e.g., placeholder values), the agent triggers a self-correction loop to verify and refine the query before finalizing the answer.

\par\medskip
\noindent\textbf{Question:} Which cities have the top 5 lowest enrollment number for students in grades 1 through 12?

\par\bigskip
\begin{minipage}{\linewidth}
\noindent\textbf{Turn 1 (Think):} I need to find the cities with the top 5 lowest enrollment numbers. I will join the tables and order by enrollment ascending.

\smallskip
\noindent\textbf{Turn 1 (SQL):}
\begin{lstlisting}[language=SQL, basicstyle=\ttfamily\small, breaklines=true, frame=single, xleftmargin=1.0em]
SELECT T2.City, T1."Enrollment (K-12)" FROM frpm AS T1 INNER JOIN schools AS T2 ON T1.CDSCode = T2.CDSCode ORDER BY T1."Enrollment (K-12)" ASC LIMIT 5;
\end{lstlisting}

\smallskip
\noindent\textbf{Turn 1 (Observation):} \texttt{[(`Markleeville', 1.0), (`Oakland', 1.0), (`Palermo', 1.0)]}
\end{minipage}

\par\bigskip
\begin{minipage}{\linewidth}
\noindent\textbf{Turn 2 (Think):} \textit{(The agent detects a potential data anomaly.)} The result shows enrollment numbers are very low (1.0), which might indicate an error in the data or the need to filter out non-enrollment entries. Let's check the data to ensure accuracy.

\smallskip
\noindent\textbf{Turn 2 (SQL):}
\begin{lstlisting}[language=SQL, basicstyle=\ttfamily\small, breaklines=true, frame=single, xleftmargin=1.0em]
SELECT CDSCode FROM frpm WHERE "Enrollment (K-12)" != 1.0 LIMIT 10;
\end{lstlisting}

\smallskip
\noindent\textbf{Turn 2 (Observation):} \texttt{[(`01100170130419'), (`01100170112607')...]}
\end{minipage}

\par\bigskip
\begin{minipage}{\linewidth}
\noindent\textbf{Turn 3 (Think):} The check confirms that there are legitimate records with other values, and '1.0' likely represents missing or placeholder data. I need to filter these out to get accurate city names.

\smallskip
\noindent\textbf{Turn 3 (SQL):}
\begin{lstlisting}[language=SQL, basicstyle=\ttfamily\small, breaklines=true, frame=single, xleftmargin=1.0em]
SELECT T2.City FROM frpm AS T1 INNER JOIN schools AS T2 ON T1.CDSCode = T2.CDSCode WHERE T1."Enrollment (K-12)" != 1.0 ORDER BY T1."Enrollment (K-12)" ASC LIMIT 5;
\end{lstlisting}

\smallskip
\noindent\textbf{Turn 3 (Observation):} \texttt{[(`Bishop'), (`Kingsburg'), (`Oakland'), (`Williams'), (`Compton')]}
\end{minipage}

\par\bigskip
\begin{minipage}{\linewidth}
\noindent\textbf{Turn 4 (Think):} Now, the cities with the lowest enrollment numbers are identified and the data looks plausible. I can provide the final SQL query.

\smallskip
\noindent\textbf{Turn 4 (Final Solution):}
\begin{lstlisting}[language=SQL, basicstyle=\ttfamily\small, breaklines=true, frame=single, xleftmargin=1.0em]
SELECT T2.City FROM frpm AS T1 INNER JOIN schools AS T2 ON T1.CDSCode = T2.CDSCode WHERE T1."Enrollment (K-12)" != 1.0 ORDER BY T1."Enrollment (K-12)" ASC LIMIT 5;
\end{lstlisting}
\end{minipage}

\subsection{Scenario 3: Overcoming Sub-optimal Logic and Boundary Errors}
\label{app:scenario_suboptimal_logic}

\interlinepenalty=5000 

This case demonstrates the agent's ability to recover from hidden logical and semantic errors, specifically regarding boundary conditions and offset calculations. Even when a query executes perfectly without syntax errors, the agent evaluates the mathematical alignment between its SQL logic and the user's intent, iteratively refining its approach to overcome sub-optimal reasoning.

\par\medskip
\noindent\textbf{Question:} What is the eligible free rate of the 10th and 11th schools with the highest enrolment for students in grades 1 through 12?

\par\bigskip
\begin{minipage}{\linewidth}
\noindent\textbf{Turn 1 (Think):} The question asks for the eligible free rate for the 10th and 11th schools based on enrollment. I need to calculate this rate, sort the schools by enrollment in descending order, and select the 10th and 11th positions. I will use the \texttt{LIMIT} clause to extract these two specific rows.

\smallskip
\noindent\textbf{Turn 1 (SQL):}
\begin{lstlisting}[language=SQL, basicstyle=\ttfamily\small, breaklines=true, frame=single, xleftmargin=1.0em]
SELECT "Free Meal Count (K-12)" / "Enrollment (K-12)" as rate FROM frpm ORDER BY "Enrollment (K-12)" DESC LIMIT 1, 2;
\end{lstlisting}

\smallskip
\noindent\textbf{Turn 1 (Observation):} \texttt{[(0.815746,), (0.338349,)]}
\end{minipage}

\par\bigskip
\begin{minipage}{\linewidth}
\noindent\textbf{Turn 2 (Think):} \textit{(The agent detects a sub-optimal logic in pagination.)} The query successfully returned results, but the logic \texttt{LIMIT 1, 2} is incorrect for finding the 10th and 11th schools. In SQL, the offset starts at 0, so an offset of 1 skips only the 1st row and returns the 2nd and 3rd rows. Furthermore, to prevent potential integer division truncation, I should cast the numerator to a real number. I need to adjust the offset to 9 to correctly target the 10th and 11th schools.

\smallskip
\noindent\textbf{Turn 2 (SQL):}
\begin{lstlisting}[language=SQL, basicstyle=\ttfamily\small, breaklines=true, frame=single, xleftmargin=1.0em]
SELECT CAST("Free Meal Count (K-12)" AS REAL) / "Enrollment (K-12)" as rate FROM frpm ORDER BY "Enrollment (K-12)" DESC LIMIT 9, 2;
\end{lstlisting}

\smallskip
\noindent\textbf{Turn 2 (Observation):} \texttt{[(0.134436,), (0.290589,)]}
\end{minipage}

\par\bigskip
\begin{minipage}{\linewidth}
\noindent\textbf{Turn 3 (Think):} The query now orders the rates by enrollment in descending order and uses the correct pagination boundary. \texttt{LIMIT 9, 2} correctly skips the top 9 schools and selects the 10th and 11th rows. The data type casting also ensures robust calculation. The results are now logical and correct.

\smallskip
\noindent\textbf{Turn 3 (Final Solution):}
\begin{lstlisting}[language=SQL, basicstyle=\ttfamily\small, breaklines=true, frame=single, xleftmargin=1.0em]
SELECT CAST("Free Meal Count (K-12)" AS REAL) / "Enrollment (K-12)" FROM frpm ORDER BY "Enrollment (K-12)" DESC LIMIT 9, 2;
\end{lstlisting}
\end{minipage}

\subsection{Scenario 4: Recovery from Schema Hallucination and Ambiguity}
\label{app:scenario_schema_hallucination}

\interlinepenalty=5000 

This case illustrates the agent's capacity to recover from schema hallucinations (e.g., mistyping or misremembering table names) and simultaneously resolve semantic ambiguities. When confronted with a database execution error, the agent not only corrects the structural typo but also proactively refines the logical filtering conditions based on a deeper understanding of the schema constraints.

\par\medskip
\noindent\textbf{Question:} Please list the zip code of all the charter schools in Fresno County Office of Education.

\par\bigskip
\begin{minipage}{\linewidth}
\noindent\textbf{Turn 1 (Think):} I need to retrieve the zip codes of all charter schools within the Fresno County Office of Education. I will join the \texttt{schools} table with the \texttt{fprm} table on the \texttt{CDSCode} and filter by the county and charter school status. 

\smallskip
\noindent\textbf{Turn 1 (SQL):}
\begin{lstlisting}[language=SQL, basicstyle=\ttfamily\small, breaklines=true, frame=single, xleftmargin=1.0em]
SELECT T1.Zip FROM schools AS T1 INNER JOIN fprm AS T2 ON T1.CDSCode = T2.CDSCode WHERE T1.County = 'Fresno County Office of Education' AND T2."Charter School (Y/N)" = 1;
\end{lstlisting}

\smallskip
\noindent\textbf{Turn 1 (Observation):} \texttt{Operational Error: no such table: fprm} 
\end{minipage}

\par\bigskip
\begin{minipage}{\linewidth}
\noindent\textbf{Turn 2 (Think):} \textit{(The agent processes the database error and reviews the schema.)} I made a typo in the table name; it should be \texttt{frpm} instead of \texttt{fprm}. Furthermore, 'Fresno County Office of Education' likely refers to the school \texttt{District} rather than the \texttt{County} itself, as the county name is typically just 'Fresno'. I will correct the table hallucination and refine the \texttt{WHERE} clause to check both \texttt{County} and \texttt{District} for precise filtering.

\smallskip
\noindent\textbf{Turn 2 (SQL):}
\begin{lstlisting}[language=SQL, basicstyle=\ttfamily\small, breaklines=true, frame=single, xleftmargin=1.0em]
SELECT T1.Zip FROM schools AS T1 INNER JOIN frpm AS T2 ON T1.CDSCode = T2.CDSCode WHERE T2."Charter School (Y/N)" = 1 AND T1.County = 'Fresno' AND T1.District = 'Fresno County Office of Education';
\end{lstlisting}

\smallskip
\noindent\textbf{Turn 2 (Observation):} \texttt{[('93726-5309',), ('93628-9602',), ('93706-2611',), ('93726-5208',), ('93706-2819',)]} 
\end{minipage}

\par\bigskip
\begin{minipage}{\linewidth}
\noindent\textbf{Turn 3 (Think):} The query executed successfully and returned a valid list of zip codes. The table name was correctly resolved, and the district/county filtering accurately identified the target schools. I can now provide the final solution.

\smallskip
\noindent\textbf{Turn 3 (Final Solution):}
\begin{lstlisting}[language=SQL, basicstyle=\ttfamily\small, breaklines=true, frame=single, xleftmargin=1.0em]
SELECT T1.Zip FROM schools AS T1 INNER JOIN frpm AS T2 ON T1.CDSCode = T2.CDSCode WHERE T2."Charter School (Y/N)" = 1 AND T1.County = 'Fresno' AND T1.District = 'Fresno County Office of Education';
\end{lstlisting}
\end{minipage}

\section{Dataset-Wide Controlled Isolation}
\label{sec:controlled_isolation}

To complement the component ablation in the main paper, we perform a dataset-wide controlled-isolation analysis on the full BIRD-dev set (1,534 queries). The goal is to quantify how many additional queries are resolved by each mechanism when starting from a strict baseline. We use the RL Generator only with $T{=}1$ and greedy decoding as the baseline configuration, which achieves 62.78\% execution accuracy.

\begin{table}[h]
    \centering
    \small
    \caption{Dataset-wide controlled isolation on BIRD-dev. Each row attributes additional solved queries beyond the strict RL-Generator baseline.}
    \label{tab:controlled_isolation_appendix}
    \begin{tabular}{lcc}
    \toprule
    \textbf{Configuration / Mechanism} & \textbf{Extra Solved Cases} & \textbf{Gain} \\
    \midrule
    Strict baseline: RL Generator only ($T{=}1$, Greedy) & --- & 62.78\% EX \\
    + Schema pruning & 56 & +3.63\% \\
    + RL interactive reasoning & 44 & +2.89\% \\
    + Generative verification / reranking & 131 & +8.54\% \\
    \midrule
    Full MARS-SQL & 231 & 77.84\% EX \\
    \bottomrule
    \end{tabular}
\end{table}

Two observations follow from Table~\ref{tab:controlled_isolation_appendix}. First, generative verification is the single largest isolated contributor, which is consistent with the strong gains observed in our selection-method analysis. Second, the RL interactive policy contributes 44 additional correct queries under greedy decoding alone, without any sampling or reranking. This directly supports our claim that the model has learned to use execution feedback for self-correction at the policy level, rather than benefiting only from test-time selection.

\section{Interaction turns impact}
\label{Interaction turns impact}
Table~\ref{tab:ablation_turns_detailed} presents an ablation study on the impact of the maximum number of interaction turns (T) allowed during training and inference. The study compares the performance of the base model against versions fine-tuned with maximum turn limits of T=1, T=5, and T=10. For each training configuration, we report three execution accuracy metrics on the BIRD-dev benchmark: Greedy (a single output), Best of 8 (the highest score among 8 candidates), and Selected (the score after reranking). The results indicate that increasing the number of turns during fine-tuning consistently improves performance. For example, the model trained with T=10 achieves the highest Greedy (69.88\%) and Best of 8 (83.89\%) accuracy when inferring with 10 turns, demonstrating the value of a larger interaction budget for complex reasoning.
\begin{table*}[t!] 
\centering 
\setlength{\tabcolsep}{4pt} \caption{Ablation study on the maximum number of interaction turns (\textit{T}). We compare the original model against models fine-tuned with different turn limits. For each setting, we report \textbf{Greedy} accuracy (from a single output), \textbf{Best of 8} accuracy, and \textbf{Selected} accuracy (after reranking 8 candidates) with grounding agent. All scores are execution accuracy (\%) on the BIRD-dev benchmark.} \label{tab:ablation_turns_detailed} 
\resizebox{\textwidth}{!}{ 
\begin{tabular}{@{}lccccccccc@{}} 
\toprule 
\textbf{\shortstack{Model \\ (Trained w/ Max Turns)}} & \multicolumn{3}{c}{\textbf{Inference (T=1)}} & \multicolumn{3}{c}{\textbf{Inference (T=5)}} & \multicolumn{3}{c}{\textbf{Inference (T=10)}} \\ \cmidrule(lr){2-4} \cmidrule(lr){5-7} \cmidrule(lr){8-10} 
& Greedy(\%) & Best of 8(\%) & Select(\%) & Greedy(\%) & Best of 8(\%) & Select(\%) & Greedy(\%) & Best of 8(\%) & Select(\%) \\ 
\midrule Original Model (Base)  & 54.43 & 77.76 & 69.69 & 55.41 & 77.82 & 70.34 & 55.76 & 77.56 & 70.07 \\Trained (T=1) & 66.41 & 78.6 & 72.06 & 66.95 & 78.76 & 72.75 & 67.60 & 80.63 & 74.19 \\ Trained (T=5) & 67.60 & 82.19 & 75.29 & 69.30 & 83.7 & 77.84 & 68.25 & 82 & 76.40 \\ Trained (T=10) & 67.73 & 83.61 & 76.86 & 69.36 & 83.95 & 77.12  & 69.88 & 83.89 & 77.57 \\ \bottomrule \end{tabular}} \end{table*}


\section{Impact of Multiple Candidate Generations}
To evaluate the impact of generating multiple candidate trajectories, we conduct a ``Best-of-N" analysis, where N is the number of parallel rollouts. As shown in Table~\ref{tab:best_of_n}, increasing the number of candidates provides a substantial performance boost. This demonstrates that the exploratory nature of our Generator agent is effective at covering the solution space, with the upper-bound performance (Pass@N) increasing consistently with more samples. The final accuracy, after applying our Generative Validation Agent, also benefits from a larger pool of high-quality candidates to select from.

\begin{table}[h!]
\centering
\caption{Impact of ``Best-of-N" selection on the BIRD-dev benchmark. \textbf{Greedy (Best of 1)} is the execution accuracy of the final selected trajectory. \textbf{Best of N} represents the upper-bound performance (Pass@N), indicating the percentage of times at least one correct trajectory was found among N candidates. Inference parameters: temperature=0.8, top\_k=50, top\_p=0.7, max\_iterations=5.}
\label{tab:best_of_n}
\begin{tabular}{@{}lc@{}}
\toprule
\textbf{Selection Strategy} & \textbf{Execution Accuracy (\%)} \\
\midrule
Greedy (Best of 1) & 69.30 \\
Best of 2 & 74.04 \\
Best of 4 & 79.71 \\
Best of 8 & 83.76 \\  
Best of 16 & 86.31 \\
Best of 32 & 87.54 \\
\bottomrule
\end{tabular}
\end{table}

\section{Table Level Grounding}
\label{grounding}
Table~\ref{tab:prompt_schema_linking} details the prompt for our RL-trained Schema Grounding Agent, which  elicit a step-by-step reasoning process during inference. It instructs the agent to analyze a given table's schema in the context of the user's question and any external knowledge. The agent is required to first articulate its analysis within `$<$think$>$' tags, followed by a final, parsable decision in `$<$answer$>$' tags. This output must specify the table's relevance (`Y'/`N') and, if applicable, a Python list of useful columns. This structured format ensures a transparent and predictable output format crucial for our framework.
\begin{table*}[t]\centering
\begin{minipage}{1.0\textwidth}\vspace{0mm}
    \caption{The prompt used to guide the agent in the table-level schema linking task. It includes the role description, task instructions, output format examples, and the prefix for the agent's response.}
    \centering
    \begin{sectionbox}[]{Prompt for Table-level Schema Linking}
    \centering
    \footnotesize
    \begin{tabular}{p{0.97\textwidth}}
    \textbf{User:}\\[2mm]
    You are doing table level schema linking. Given a table with schema information and the task, you should think step by step and decide whether this table is related to the task. \\
    Your thought process should be enclosed in \texttt{<think></think>} tags, and your final decision in \texttt{<answer></answer>} tags. \\
    For the answer, first state `Y' for relevant or `N' for not relevant. If relevant, also provide a Python list of the column names you believe are most useful. \\[3mm]
    Example of a final answer format: \\
    \texttt{<answer>}\\
    \texttt{Y}\\
    \texttt{["player\_name", "team\_name", "matches\_played"]}\\
    \texttt{</answer>}\\[3mm]
    or\\[3mm]
    \texttt{<answer>}\\
    \texttt{N}\\
    \texttt{</answer>}\\[3mm]
    Here is the information for the current task:\\[3mm]
    \textbf{\#\#\# Table Information:} \\
    \textit{\{table\_info\}} \\
    \textbf{\#\#\# User Question:} \\
    \textit{\{task\}} \\
    \textbf{\#\#\# External Knowledge (if any):} \\
    \textit{\{external\}} \\[5mm]
    \hline \\[-1.5mm]
    \textbf{Assistant:}\\[2mm]
    Let me solve this step by step. \\
    \texttt{<think>} \\
    \end{tabular}
    \end{sectionbox}
    \vspace{-2mm}
    \label{tab:prompt_schema_linking}
\end{minipage}
\end{table*}

Table~\ref{tab:ablation_grounding} presents recall and precision statistics for our schema grounding agent, comparing our RL-based approach against the base model and a version trained with Supervised Fine-Tuning (SFT). The results clearly demonstrate the superiority of our method, which achieves exceptionally high recall and precision across all benchmarks. On the complex in-domain BIRD-dev set, our primary concern is recall. Our agent achieves a recall of 97.78\%, with only 48 examples failing to identify all required schema components, which we consider a highly effective result. Simultaneously, it maintains a high precision of 90.74\%, indicating that the selections are not only comprehensive but also accurate. This strong performance extends to the out-of-domain Spider-test and Spider-DK benchmarks, underscoring the robustness of our RL-trained grounding agent.
\begin{table*}[t!]
\centering
\caption{Recall and precision statistics after grounding for Bird-dev, Spider-test and Spider-DK. \textbf{Recall} measures the percentage of instances where all required columns were identified. \textbf{Precision} measures the ratio of required columns to all selected columns, indicating the selection's accuracy.}
\label{tab:ablation_grounding}
\begin{tabular}{@{}lcccccc@{}}
\toprule
\textbf{Grounding Model} & \multicolumn{2}{c}{\textbf{Bird dev}} & \multicolumn{2}{c}{\textbf{Spider test}} & \multicolumn{2}{c}{\textbf{Spider DK}} \\
\cmidrule(lr){2-3} \cmidrule(lr){4-5} \cmidrule(lr){6-7}
& Recall (\%) & Precision (\%) & Recall (\%) & Precision (\%) & Recall (\%) & Precision (\%) \\
\midrule
Qwen 7B (Base) & 68.59 & 53.45 & 87.48 & 69.22 & 84.25 & 66.54 \\
Qwen 7B + SFT & 74.97 & 67.01 & 90.39 & 78.16 & 88.60 & 72.71 \\
\textbf{Qwen 7B + RL (Ours)} & 97.78 & 90.74 & 98.97 & 93.62 & 98.13 & 91.59 \\
\bottomrule
\end{tabular}
\end{table*}

The practical impact of the Grounding Agent depends on the schema regime. On smaller and cleaner schemas such as Spider, the Generator can often operate with comparatively limited interference, so the end-to-end gain from grounding is smaller. On BIRD-dev, however, the search space is much larger, making schema filtering substantially more valuable. More broadly, this requirement to preserve salient evidence while filtering large amounts of irrelevant context is also consistent with observations from long-form and long-context analysis~\citep{he2026stablelongformgenerationbenchmarking}. Table~\ref{tab:grounding_dataset_stats} summarizes this contrast.

\begin{table}[h]
    \centering
    \small
    \caption{Dataset characteristics relevant to the Grounding Agent. Larger schemas create more distractors for downstream generation.}
    \label{tab:grounding_dataset_stats}
    \begin{tabular}{lccc}
    \toprule
    \textbf{Dataset} & \textbf{Avg. Schema Tokens} & \textbf{Avg. Columns/DB} & \textbf{Avg. Tokens/SQL} \\
    \midrule
    Spider & $\sim$250 & 26.82 & 30.9 \\
    BIRD-dev & $\sim$950 & 54.56 & 18.5 \\
    \bottomrule
    \end{tabular}
\end{table}

This context also helps explain why the Grounder is more important under semantic ambiguity and domain shift. Spider-DK is specifically designed to break clean lexical matches between questions and schemas, and Table~\ref{tab:ablation_grounding} shows that our RL Grounder maintains very high recall and precision there as well. In this sense, grounding is best understood not as a uniformly large gain on every benchmark, but as a component whose importance increases with schema scale, ambiguity, and cross-domain difficulty.

\section{Generation Agent Prompt}
Table~\ref{tab:prompt_sql_reasoning_example} details the comprehensive prompt structure used to guide the agent's multi-turn generation process. The prompt establishes the agent's persona as a data science expert and provides all necessary context, including the database schema, external knowledge, and the user's question. It strictly enforces an output format that requires the agent to vocalize its reasoning within $<$think$>$ blocks before executing a query in a $<$sql$>$ block. The database returns feedback in an $<$observation$>$ block, which the agent uses for subsequent reasoning turns, ultimately providing the final answer in a $<$solution$>$ block. This iterative structure is designed to facilitate a dynamic, step-by-step problem-solving process.

\begin{table*}[ht!]\centering
\begin{minipage}{1.0\textwidth}\vspace{0mm}    \centering
\begin{sectionbox}[]{Prompt Format for SQL Reasoning}
    \centering
      \footnotesize
    \begin{tabular}{p{0.97\textwidth}}
\textbf{Prompt Description:} \\
You are a data science expert. Below, you are provided with a database schema and a natural language question. Your task is to understand the schema and generate a valid SQL query to answer the question within limited turns. You should breakdown the problem, draft your reasoning process, and generate the solution. \\

\textbf{Database Engine:} \\
SQLite \\

\textbf{Database Schema:} \\
\{db\_details\} \\
This schema describes the database's structure, including tables, columns, primary keys, foreign keys, and any relevant relationships or constraints. \\

\textbf{External Knowledge:} \\
\{external\_knowledge\} \\

\textbf{Question:} \\
\{question\} \\

\textbf{Important Instructions:} \\
- Make sure you only output the information that is asked in the question. If the question asks for a specific column, make sure to only include that column in the SELECT clause, nothing more. \\
- The generated query should return all of the information asked in the question without any missing or extra information. \\
- Before generating the final SQL query, please think how to write the query. It should include detailed considerations such as analysing questions, summarizing relevant findings, brainstorming new ideas, verifying the accuracy of the current steps, refining any errors, thinking of how to call SQL tools, and revisiting previous steps. \\

\textbf{Output Format (STRICTLY ENFORCED):} \\
- Conduct thinking inside \texttt{<think>...</think>} blocks every time you get new observation or information. Start with \texttt{<think>...</think>} blocks in your responses as shown in the following example. \\
- You can use SQL tool written within a single \texttt{<SQL>your SQL</SQL>} block to explore or verify. You can't use the format \texttt{```SQL ; \textbackslash n```}, you must use the format \texttt{<SQL>your SQL</SQL>} to get the output. \texttt{<SQL>your SQL</SQL>} block should follow closely behind \texttt{<think>...</think>} block. SQL tool output will be shown as dataframe inside \texttt{<observation>...</observation>}. Based on this observation, you can think again and refine. \\
- The returned dataframe will be truncated in 50 rows if observation is too long. \\
- If you find no further exploration is needed or have only 1 turn left, you MUST directly provide the final SQL query solution inside \texttt{<solution>...</solution>}. \\
- All your responses should be in the \texttt{<think>...</think>}, \texttt{<sql>...</sql>}, \texttt{<observation>...</observation>}, \texttt{<solution>...</solution>} blocks. \\

\vspace{5pt}
\textbf{Example:} \\
\textbf{Question:} how many pigs are in the farm? \\
\textbf{Database Schema:} \\
Table: animals \\
- id (INTEGER, PRIMARY KEY) \\
- species (TEXT) \\
- age (INTEGER) \\
- name (TEXT) \\

\textbf{Output:} \\
\texttt{<think>I am querying how many pigs are in the farm. I will begin by checking if the 'animals' table exists and contains entries with species = 'pig'.</think>} \\
\texttt{<SQL>SELECT COUNT(*) FROM animals WHERE species = 'pig';</SQL>} \\
\texttt{<observation>} \\
\texttt{+----------+} \\
\texttt{| COUNT(*) |} \\
\texttt{+----------+} \\
\texttt{|   12     |} \\
\texttt{+----------+} \\
\texttt{</observation>} \\
\texttt{<think>The result indicates that there are 12 pigs in the farm. Since the question asks for how many pigs, I can now output the final SQL as the solution.</think>} \\
\texttt{<solution>SELECT COUNT(*) FROM animals WHERE species = 'pig';</solution>} \\

    \end{tabular}
\end{sectionbox}
\vspace{-2mm}
\caption{Prompt example for SQL reasoning without memory.}
\label{tab:prompt_sql_reasoning_example}
\end{minipage}
\end{table*}

\section{Validation Agent}
\label{verifier}
Our Generative Validation Agent is guided by the prompt detailed in Table~\ref{tab:prompt_verifier_updated}. The prompt instructs the agent to act as an expert SQL data analyst, with the objective of evaluating the logical correctness of a proposed SQL solution for a given problem. Unlike our previous approach, this prompt no longer constrains the agent to reason about a sampled or truncated database. Instead, it assumes the agent evaluates the query's validity against the full database schema and context. The prompt structure provides the agent with the user's question, the candidate SQL solution, and a dedicated field for any relevant ``External Knowledge" that might be necessary for a correct evaluation. The output format remains strict, requiring the agent to begin its response with a definitive ``Yes" or ``No" before any subsequent reasoning.
\begin{table*}[t!]
\centering
\begin{minipage}{1.0\textwidth}\vspace{0mm}
    \caption{The prompt used for the Generative Validation Agent. The agent is framed as a SQL expert and is provided with the problem, the proposed SQL query, and any relevant external knowledge. It evaluates the logical correctness of the query and must provide a final ``Yes" or ``No" judgment.}
    \centering
    \begin{sectionbox}[]{Prompt for Generative Validation Agent}
    \centering
    \footnotesize
    \begin{tabular}{p{0.97\textwidth}}
    \textbf{User:}\\[3mm]
    \textbf{Task Background:}\\
    You are an expert SQL data analyst. Your task is to verify if a proposed solution correctly answers a user's question.\\[3mm]

    \textbf{Problem:}\\
    \textit{\{question\}}\\[3mm]

    \textbf{External Knowledge:}\\
    \textit{\{external\_knowledge\}}\\[3mm]

    \textbf{Proposed Solution:}\\
    \textit{\{solution\_text\}}\\[3mm]

    \rule{\textwidth}{0.4pt}\\[1mm] 

    \textbf{Your Task:}\\
    Based on all the information, is the SQL query in the solution logically correct for answering the question?\\
    You must answer with ``Yes" or ``No" first, before any other text.\\[3mm]
    
    Is the answer correct (Yes/No)? \\
    \end{tabular}
    \end{sectionbox}
    \vspace{-2mm}
    \label{tab:prompt_verifier_updated}
\end{minipage}
\end{table*}

\section{LLM as a judge prompt}
The prompt in Table~\ref{tab:prompt_selection_agent} is used for our baseline selection method LLM as a judge. This prompt is designed to guide the model in identifying the optimal SQL query from a set of generated candidates. The agent is explicitly instructed to consider each candidate's associated reasoning, the SQL query itself, and most crucially, its execution observation on the database. This emphasis on execution results is paramount, as it allows the agent to distinguish between syntactically correct queries and those that truly provide the correct and complete answer to the user's question, even if a query might appear correct but yields erroneous or empty results. After presenting the user's question and the formatted candidate solutions (each including reasoning, SQL, and execution output), the prompt concludes with strict instructions for the agent to output only the index number of the single best candidate. In cases of ties, the candidate with the lowest index is to be chosen, ensuring a deterministic selection process.
\begin{table*}[h!]
\centering
\begin{minipage}{1.0\textwidth}\vspace{0mm}
    \caption{The prompt used for the Selection Agent, operating as an LLM judge. It guides the model to select the best SQL query from multiple candidates by evaluating their reasoning, SQL code, and critically, their execution observations. Strict output instructions ensure a direct index selection.}
    \centering
    \begin{sectionbox}[]{Prompt for Selection Agent (LLM as a Judge)}
    \centering
    \footnotesize
    \begin{tabular}{p{0.97\textwidth}}
    \textbf{User:}\\[3mm]
    \textbf{Task Background:}\\
    You are an expert SQL data analyst. Your task is to select the BEST SQL query that correctly answers a user's question.\\[3mm]

    You are given several candidates. For each candidate, you will see its reasoning, the SQL query itself, and importantly, \textbf{the result of executing that query on the database.} A query might look correct but return an error or empty/wrong data. You must use the execution observation to make your final decision.\\[3mm]

    Here is the user's question:\\
    \textit{\{question\}}\\[3mm]

    Evaluate the following candidates based on ALL available information. Does the ``Execution Observation" for a candidate actually answer the user's question?\\
    ---\\[1mm]
    \textit{\{formatted\_candidates\}}\\[1mm]
    ---\\[3mm]

    \textbf{Final Analysis:}\\
    Considering the reasoning, the SQL code, and especially the \textbf{execution results}, which single candidate provides the most correct and complete answer to the user's question?\\[3mm]

    \textbf{Instructions for your response:}\\
    \begin{itemize}
        \item Respond with ONLY the index number of the single best candidate.
        \item If multiple candidates produce correct results, select the one with the LOWEST index number.
        \item Do not include any other words, symbols, or explanations.
    \end{itemize}\\[3mm]

    Best candidate index:\\
    \end{tabular}
    \end{sectionbox}
    \vspace{-2mm}
    \label{tab:prompt_selection_agent}
\end{minipage}
\end{table*}

\section{Selection method comparison}
\label{Selection method comparison}
We compare our proposed Generative Verifier against several strong baselines for trajectory selection, with the results detailed in Table~\ref{tab:ablation_verifier}. The initial Pass@8 accuracy of our Generator agent's output establishes the theoretical upper bound for any selection method, as it represents the percentage of questions for which at least one of the eight generated trajectories is correct.
\begin{table*}[ht]
\centering
\caption{Ablation study of different selection strategies. The first row, \textbf{Pass@8}, shows the baseline execution accuracy (\%) of the eight candidate trajectories from our Generator agent before any selection. Subsequent rows report the final accuracy after applying each method to select the best trajectory. \textbf{Self-Consistency} picks the most frequent result, \textbf{LLM as a Judge} uses GPT-4o/Qwen for selection, and \textbf{Ours} uses our fine-tuned 7B Generative Verifier.}
\label{tab:ablation_verifier}
\begin{tabular}{@{}llccc@{}}
\toprule
\textbf{Method} & \textbf{Model Size} & \textbf{Bird dev (\%)} & \textbf{Spider test (\%)} & \textbf{Spider DK (\%)} \\
\midrule
\textit{Pass@8 (Generator Output)} & - & \textit{83.76} & \textit{90.68} & \textit{82.06} \\
\midrule
LLM as a Judge (GPT-4.1) & Unkonwn & 75.15 & 83.47 & 71.40 \\
LLM as a Judge (Qwen) & 7B & 70.47 & 79.60 & 70.09 \\
Self-Consistency & - & 72.93 & 83.51 & 73.08 \\
\textbf{Ours (Generative Verifier)} & \textbf{7B} & \textbf{77.84} & \textbf{89.75} & \textbf{78.13} \\
\bottomrule
\end{tabular}
\end{table*}

\end{document}